\documentclass[11pt]{article}
\usepackage[table]{xcolor} 
\usepackage[final]{acl}

\usepackage{times}
\usepackage{latexsym}
\usepackage{listings} 

\usepackage[T1]{fontenc}

\usepackage[utf8]{inputenc}

\usepackage{microtype}

\usepackage{inconsolata}

\usepackage{graphicx}
\usepackage[linesnumbered,ruled,vlined]{algorithm2e}
\usepackage{amsmath}
\SetKwInput{KwRequire}{Require}
\usepackage{tabularx}
\usepackage{float}
\usepackage{hhline}
\usepackage{array}
\usepackage{multirow}
\usepackage{makecell} 
\usepackage{tipa} 
\usepackage{afterpage}
\usepackage[normalem]{ulem}
\usepackage{longtable}
\usepackage{booktabs}

\title{Culture-TRIP: Culturally-Aware Text-to-Image Generation 

with Iterative Prompt Refinement}

\author{
  \textbf{Suchae Jeong\textsuperscript{1}}\thanks{indicates equal contribution.},
  \textbf{Inseong Choi\textsuperscript{1}}\footnotemark[1],
  \textbf{Youngsik Yun\textsuperscript{2}},
  \textbf{Jihie Kim\textsuperscript{2}\thanks{indicates corresponding authors.}},
\\
\\
  \textsuperscript{1}Department of Computer Science and Engineering, Dongguk University,\\
  \textsuperscript{2}Department of Computer Science and Artificial Intelligence, Dongguk University
\\
  \small{
    \href{mailto:email@domain}{yys3606@dgu.ac.kr}, \href{mailto:email@domain}{jihie.kim@dgu.edu}
  }
}

\begin{document}
\maketitle
\begin{abstract}
Text-to-Image models, including Stable Diffusion, have significantly improved in generating images that are highly semantically aligned with the given prompts. However, existing models may fail to produce appropriate images for the cultural concepts or objects that are not well known or underrepresented in western cultures, such as `hangari' (Korean utensil). In this paper, we propose a novel approach, \textbf{Culturally-Aware Text-to-Image Generation with Iterative Prompt Refinement (Culture-TRIP)}, which refines the prompt in order to improve the alignment of the image with such culture nouns in text-to-image models. Our approach (1) retrieves cultural contexts and visual details related to the culture nouns in the prompt and (2) iteratively refines and evaluates the prompt based on a set of cultural criteria and large language models. The refinement process utilizes the information retrieved from Wikipedia and the Web. Our user survey, conducted with 66 participants from eight different countries demonstrates that our proposed approach enhances the alignment between the images and the prompts. In particular, C-TRIP demonstrates improved alignment between the generated images and underrepresented culture nouns. Resource can be found at https://shane3606.github.io/Culture-TRIP.
\end{abstract}

\section{Introduction}
\begin{figure}
    \centering
    \includegraphics[width=1\linewidth]{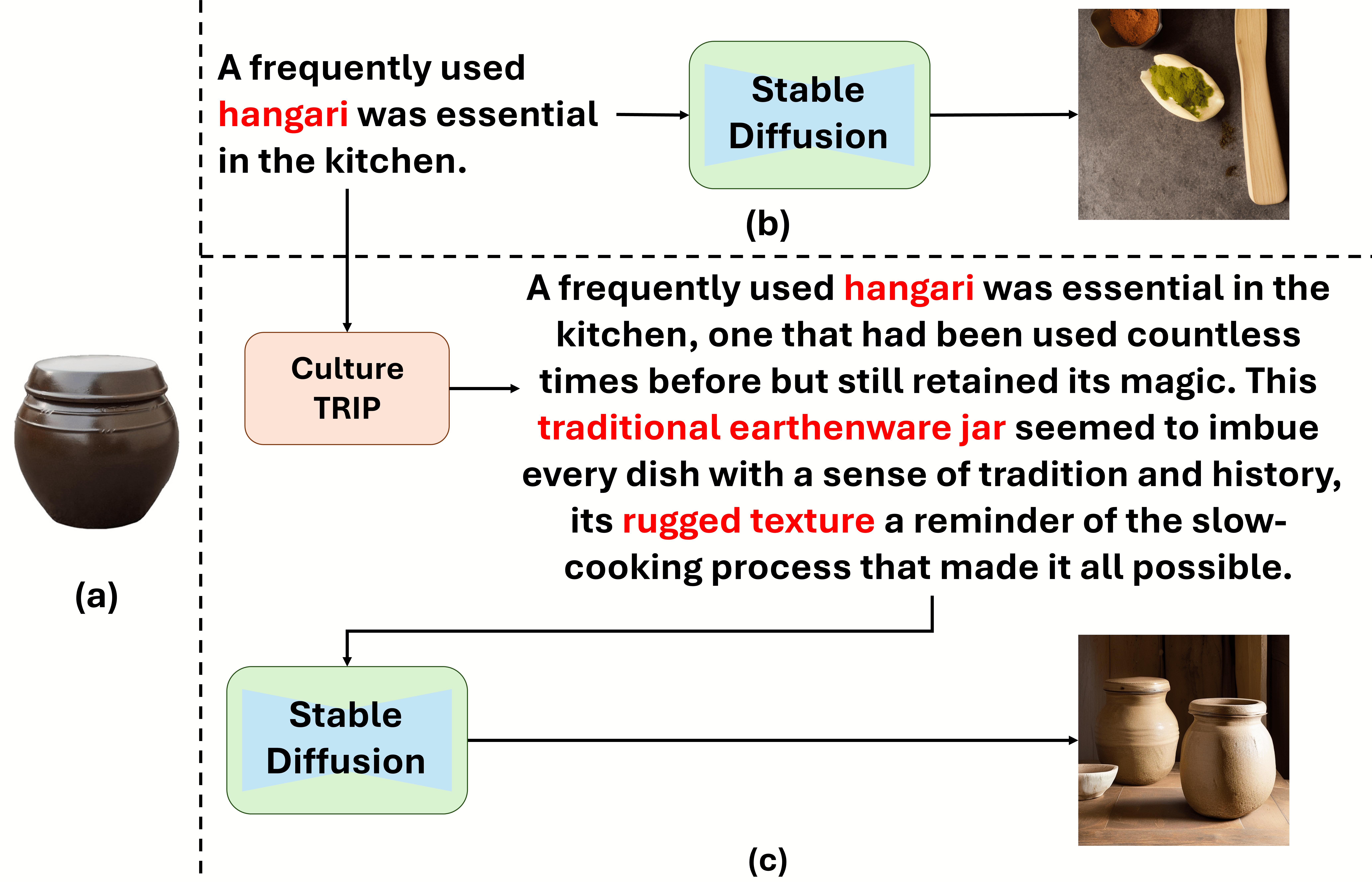}
    \caption{Comparison between Stable Diffusion with and without our proposed approach, C-TRIP. (a) shows an image of a \textit{hangari} from Wikipedia. (b) is an image generated by Stable Diffusion 2, while (c) shows an image generated with our approach. The additional knowledge about \textit{hangari} (highlighted in red) in (c) helps the model generate an image that more closely resembles the actual \textit{hangari}.}
    \label{Figure1}
\end{figure}

To date, many Text-to-Image Models ~\citep{ramesh2022hierarchical,rombach2022high,ruiz2023dreambooth} have demonstrated remarkable improvements. Despite the outstanding performance in the text-to-image models, the models fail to align the images with the culture nouns in the prompts, such as `ao dai’(a Vietnamese clothing) or `hangari’ (a Korean utensil). Most of these issues stem from the large training datasets gathered by crawling the Internet without paying attention to the details of the cultural elements ~\citep{yun2024cic}. Furthermore, Internet access varies significantly across countries ~\citep{birhane2023hate,luccioni2024stable}, leading to challenges in appropriately aligning culture nouns due to insufficient data, as shown in Figure \ref{Figure1} (b).

Representation is important in AI applications. Appropriate representation can positively affect viewers, while inappropriate ones can negatively affect them and can even be harmful ~\citep{castaneda2018power}. Culture nouns are essential elements that often represent the identity and the uniqueness of the given culture. The misrepresentation of culture nouns by existing text-to-image models may cause dissatisfaction in the corresponding countries. Moreover, the models may reinforce harmful stereotypes about particular cultures.

In this paper, we introduce a new approach that generates more culturally aligned images for the given culture nouns, called \textbf{Culturally-Aware Text-to-Image Generation with Iterative Prompt Refinement (C-TRIP)}. C-TRIP focuses on refining the prompt to ensure that the culture nouns are appropriately represented in the generated images, as shown in Figure \ref{Figure1} (c). Our goal is to generate culturally-aware images by appropriately aligning culture nouns with images from underrepresented countries. 

Our research question is, \textit{How can we refine the prompt so that text-to-image models generate images that appropriately align with culture nouns?} The Culture Capsules ~\citep{taylor1961culture} is an educational approach designed to help learners who have not directly experienced a culture gain the proper understanding. This approach explains cultural contexts and visual details, enabling learners to gain comprehensive understand unfamiliar cultures. Through this process, learners can develop a deeper profound awareness of different cultures.

Inspired by the Culture Capsules, C-TRIP first retrieves cultural contexts and visual details related to the culture nouns in the prompt and then iteratively refines and evaluates the prompt against the criteria used in culture education. Large Language Models (LLMs) are used for this iterative refinement and evaluation process, guiding the text-to-image model in generating images for culture nouns.

We conducted experiments across eight countries, refining a total of 10,000 prompts by using 50 prompt templates for each of the 25 culture nouns per country. To evaluate our results, we recruited participants who were a native of the corresponding country and had a high familiarity with the culture to rank images generated by Stable Diffusion 2 ~\citep{rombach2022high}, with or without the proposed iterative prompt refinement. The 66 participants across eight countries evaluated the 990 generated images, and C-TRIP received ratings for cultural alignment that were 18.84\% on average higher than the baseline's. In particular, our approach demonstrated more improvement for relatively the Unrecognized/Underrepresented Culture nouns (\textit{UC nouns} for short) than for the Recognized/Common Culture nouns (\textit{RC nouns} for short).

Our contributions are as follows:

\begin{enumerate}
    \item We introduce culturally-aware image generation with a prompt that improves the representation of `culture nouns'—cultural concepts or objects often overlooked by existing text-to-image models.
    \item We propose a novel approach, \textbf{C-TRIP} (\textbf{C}ulturally-Aware \textbf{T}ext-to-Image Gene\textbf{R}ation with \textbf{I}terative \textbf{P}rompt Refinement), which iteratively refines the prompt in order to improve the alignment of culture nouns in the images generated by text-to-image models.
    \item Human evaluations by representatives across eight countries demonstrate that our refined prompts enhance the alignment between generated images and culture nouns. In particular, C-TRIP demonstrated more improvement for UC nouns than RC nouns.
\end{enumerate}

\begin{figure*}
    \centering
    \includegraphics[width=\textwidth]{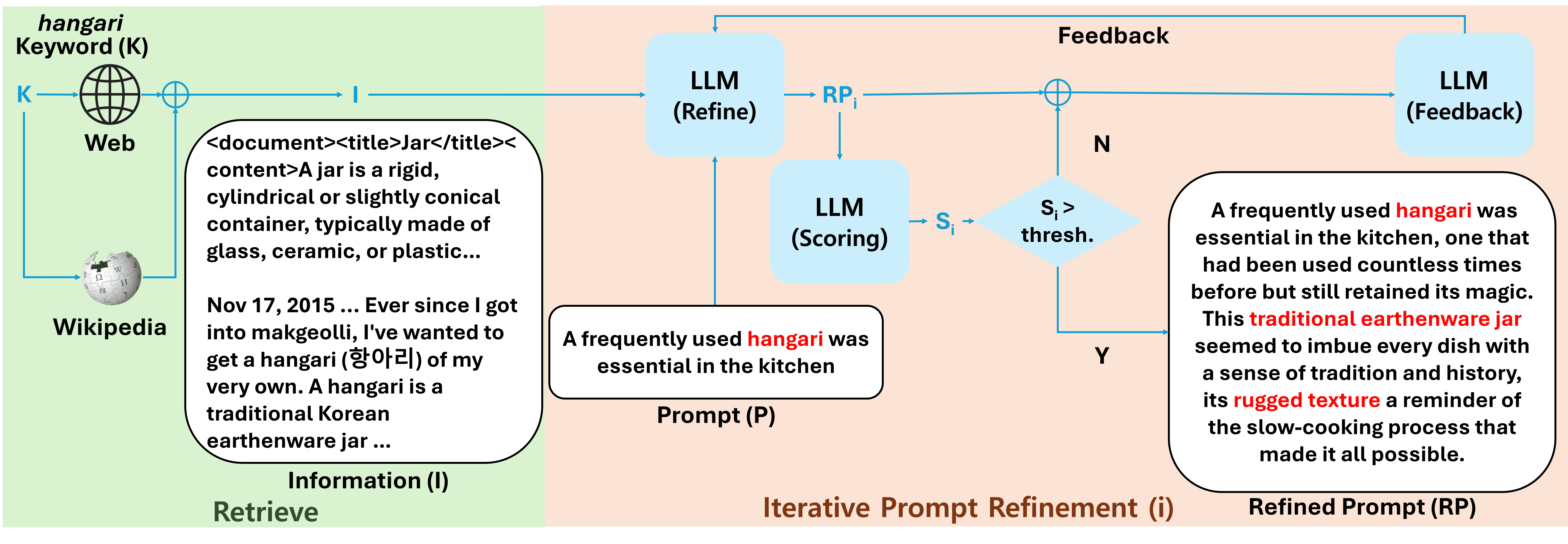}
    \caption{C-TRIP Overview. First, retrieve cultural contexts (cultural background, purpose) and visual details related to the culture nouns as described in Section \ref{sec:retrieve}. Then, refining the prompt based on the obtained information. We iteratively evaluate and refine the prompt as described in Section \ref{sec:refine}.}
    \label{Figure2}
\end{figure*}

\section{Related Work}
\subsection{Cultural Text-to-Image Generation}
Previously, various methods have been proposed to address the cultural bias in text-to-image models ~\citep{cho2023dall,friedrich2023fair,luccioni2024stable}. \citet{liu2024scoft} collected a cultural dataset called CCUB across nine cultural categories for five countries and proposed a training technique named SCoFT to address cultural bias. Similarly, \citet{kannen2024beyond} collected the CUBE across three cultural categories for eight countries. However, these approaches are highly resource-intensive, demanding significant time and cost for data collection.

Other works ~\citep{basu2023inspecting,bansal2022well} attempted to mitigate cultural bias by modifying prompt. However, merely adding contextual information such as country names to prompt is proven insufficient in mitigating cultural bias, particularly for the concepts from underrepresented countries.

Unlike the previous approaches, our approach refines prompt based on cultural information and visual details to improve the alignment of text-to-image models with culture nouns, which are significant for representing unique concepts and objects across cultures.

\subsection{Prompt Engineering in Text-to-Image Generation}
The prompt as input to the text-to-image generation guides the images created by the models. Many studies ~\citep{oppenlaender2023taxonomy,liu2022design,brade2023promptify} on prompt engineering aimed at optimizing user-desired images by text-to-image models.

\citet{wen2024hard} propose a method for learning hard prompts that is optimized for text-to-image generation. This approach demonstrates how to generate prompt with minimal tokens that effectively guide the model to produce images in a specific style. \citet{yao2024promptcot} propose a refinement method that aligns input prompt with training prompts, ensuring that text-to-image models produces high-quality images.

 Unlike the existing approaches, we refine the prompt based on cultural information related to culture nouns, guiding text-to-image models to generate images that align appropriately with cultural representation.

\subsection{Refinement}
Refinement is a method designed to improve output quality through feedback and the application of a refiner. Learned Refiners ~\citep{schick2022peer,saunders2022self} involve providing feedback through a trained refiner, which requires human-annotated data for the training process. In contrast, Prompted Refiners ~\citep{peng2023check,yang2022re3} offer feedback through prompting without the trained refiner. Recently, a method called self-refine ~\citep{madaan2024self} has been proposed, which performs iterative feedback and refinement using a single Large Language Model (LLM) without external supervision.

Unlike the self-refine methods designed for LLM tasks, we iteratively refine the prompt based on cultural contexts and visual details to enhance the understanding of culture nouns in text-to-image models.

\section{Method}
Inspired by the Culture Capsules ~\citep{taylor1961culture}, an educational method that introduces unfamiliar cultures through cultural contexts and visual details, our method refines the prompt for text-to-image models by incorporating cultural contexts and visual details relevant to the culture noun. In order to include only information essential for cultural expressions in the image, external information is retrieved, and an iterative refinement and evaluation process is conducted based on the criteria derived from both cultural contexts and visual details.

Our overall architecture is depicted in Figure \ref{Figure2}. We first retrieve cultural information from Wikipedia and Web content (Section \ref{sec:retrieve}), then second iteratively refine and evaluate the prompt based on scores (Section \ref{sec:refine}) until the stop condition is satisfied.

\subsection{Cultural Information Retrieval}\label{sec:retrieve}
In obtaining raw information related to culture nouns, we use two sources. Given a culture noun, we first retrieve the Wikipedia content in order to leverage cultural contexts and visual details. Second, for certain culture nouns that lack sufficient information on Wikipedia, we perform additional retrieval from the Web. By making use of both Wikipedia and the Web, we can collect sufficient raw data, even for relatively uncommon UC nouns.

\begin{table*}[]
\resizebox{\linewidth}{!}{%
\centering
\begin{tabular}{ccl}
\toprule
Aspect                           & Criterion                   & \multicolumn{1}{c}{Description} \\ \midrule
\multirow{3}*[-3.1ex]{Cultural Contexts} & \textbf{Clarity}            & \begin{tabular}[c]{@{}l@{}}The overall clarity of the information, specifically whether\\ the necessary details to explain the culture noun\\ are clearly and easily conveyed.\end{tabular} \\ \cmidrule(l){2-3} 
                                   & \textbf{Background}         & \begin{tabular}[c]{@{}l@{}}Whether the prompt provides appropriate historical or\\ temporal context.\end{tabular} \\ \cmidrule(l){2-3} 
                                   & \textbf{Purpose}            & \begin{tabular}[c]{@{}l@{}}Whether the prompt describes the purpose or usage of the\\ culture noun.\end{tabular} \\ \midrule
\multirow{2}*[-3.1ex]{Visual Details}    & \textbf{Visual Elements}    & \begin{tabular}[c]{@{}l@{}}Whether sufficient visual information, such as color and\\ shape, is provided.\end{tabular} \\ \cmidrule(l){2-3} 
                                   & \textbf{Comparable Objects} & \begin{tabular}[c]{@{}l@{}}Whether the prompt offers a well-known or famous ex-\\ ample by drawing a comparison to the culture nouns.\end{tabular} \\ \bottomrule
\end{tabular}
}
\caption{Criteria for scoring refined information. C-TRIP performs scoring of the refined prompt based on five criteria across the aspects of cultural contexts and visual details. Cultural contexts are a criterion for evaluating cultural information, and visual details are criteria for assessing visual information relevant to image generation. Each criterion is assigned a score ranging from 0 to 10.}
\label{Table1}
\end{table*}

\subsection{Iterative Prompt Refinement}\label{sec:refine}
The iterative refining process consists of 3 key steps: \textit{Refine}, \textit{Scoring}, and \textit{Feedback}. The \textit{Refine} step refines the raw information using the prompt and feedback from the previous \textit{Feedback} step. The \textit{Scoring} step evaluates the refined prompt based on five cultural criteria. Finally, the \textit{Feedback} step proposes revisions in the prompt based on the refined prompt and the evaluation score. 
All three steps were implemented using LLaMA-3-70B~\citep{dubey2024llama}, as iterative refinement processes benefit from larger LLM~\citep{madaan2024self} with the latest open-source model.

\paragraph{Refine}
The retrieved raw information \textit{\textbf{I}} for the culture noun \textit{\textbf{K}} is used to refine the prompts in the \textit{Refine} step. In this step, the refined prompt \textit{\textbf{RP}} is typically generated based on feedback \textit{\textbf{F}}. For the first step, only the raw information \textbf{\textit{I}} and the prompt \textbf{\textit{P}} are used. In the equations, $\parallel$ denotes concatenation throughout the paper.

\begin{equation}
\small
\hspace{-1em} 
RP_i = 
\begin{cases} 
    Refine(K || I || P) & \text{\scriptsize if $i = 0$}, \\
    Refine(K || I || RP_{i-1} || F_{i-1}) & \text{\scriptsize if $i > 0$}.
\end{cases}
\label{refine}
\end{equation}

Through the \textit{Refine} step, only the information essential for cultural education is extracted from the raw data. The final refined prompt guides the text-to-image model in generating images for culture nouns. 

\paragraph{Scoring}
In the \textit{Scoring} step, the refined prompts are evaluated based on five cultural criteria: Clarity, Background, Purpose, Visual Elements, and Comparable Objects. If the total score does exceed the threshold (thresh.) or a specified maximum iteration \textit{i}, the process stops; otherwise, it proceeds to the \textit{Feedback} step for further refinement.

\begin{equation}
score_i = Scoring(K || RP_{i})
\label{scoring}
\end{equation}

The five scoring criteria are structured based on the Culture Capsules ~\citep{taylor1961culture}, which organize the criteria into two primary aspects: cultural contexts and visual details. Specifically, \textit{Clarity}, \textit{Background}, and \textit{Purpose} are categorized as cultural contexts, while \textit{Visual Elements} and \textit{Comparable Objects} fall under visual details. Detailed descriptions of each evaluation criterion are provided in Table \ref{Table1}.
 
\paragraph{Feedback}
In the \textit{Feedback} step, each score is reviewed based on the criteria, and feedback is provided along with suggestions to improve the scores. 

\begin{equation}
F_i = Feedback(K || RP_i || score_i)
\label{feedback}
\end{equation}

Details illustrating the evolution of prompts during the refinement process, along with templates for each step, can be found in Appendix \ref{sec:Iterative Prompt Refinement}

\section{Experiments}
\subsection{Data Preparation}
\paragraph{Culture Nouns} The culture nouns are phonetically transcribed into English based on the original pronunciation in their respective languages (e.g., hangari, pronounced /\textipa{hA:NgA:ri}/). However, when an English equivalent exists, they are represented in the format of `Adjective form for the country + English expression' (e.g., Korean pagoda) to signify culture nouns.

\paragraph{Setup}Based on ~\citet{basu2023inspecting}, which addresses cultural bias in text-to-image generation by modifying prompts, we selected eight countries representing diverse cultural backgrounds: India, Pakistan, China, Japan, South Korea, Vietnam, the United States, and Germany. Drawing on research by ~\citet{liu2024scoft} on culturally-aware text-to-image models, we focused on eight specific categories that typically represent culture in visual expressions: architecture, city landmarks, clothing, dance \& music, visual arts, food \& drink, religion \& festivals, and utensils and tools. We generated 10,000 prompts using 50 prompt templates for each of the 25 culture nouns per country. These templates, generated by GPT-4o, incorporate culture nouns into typical scenarios, enabling consistent prompt generation for experimentation without relying on real-world prompts.

\begin{figure*}[!htb]
    \centering
    \includegraphics[width=1\textwidth]{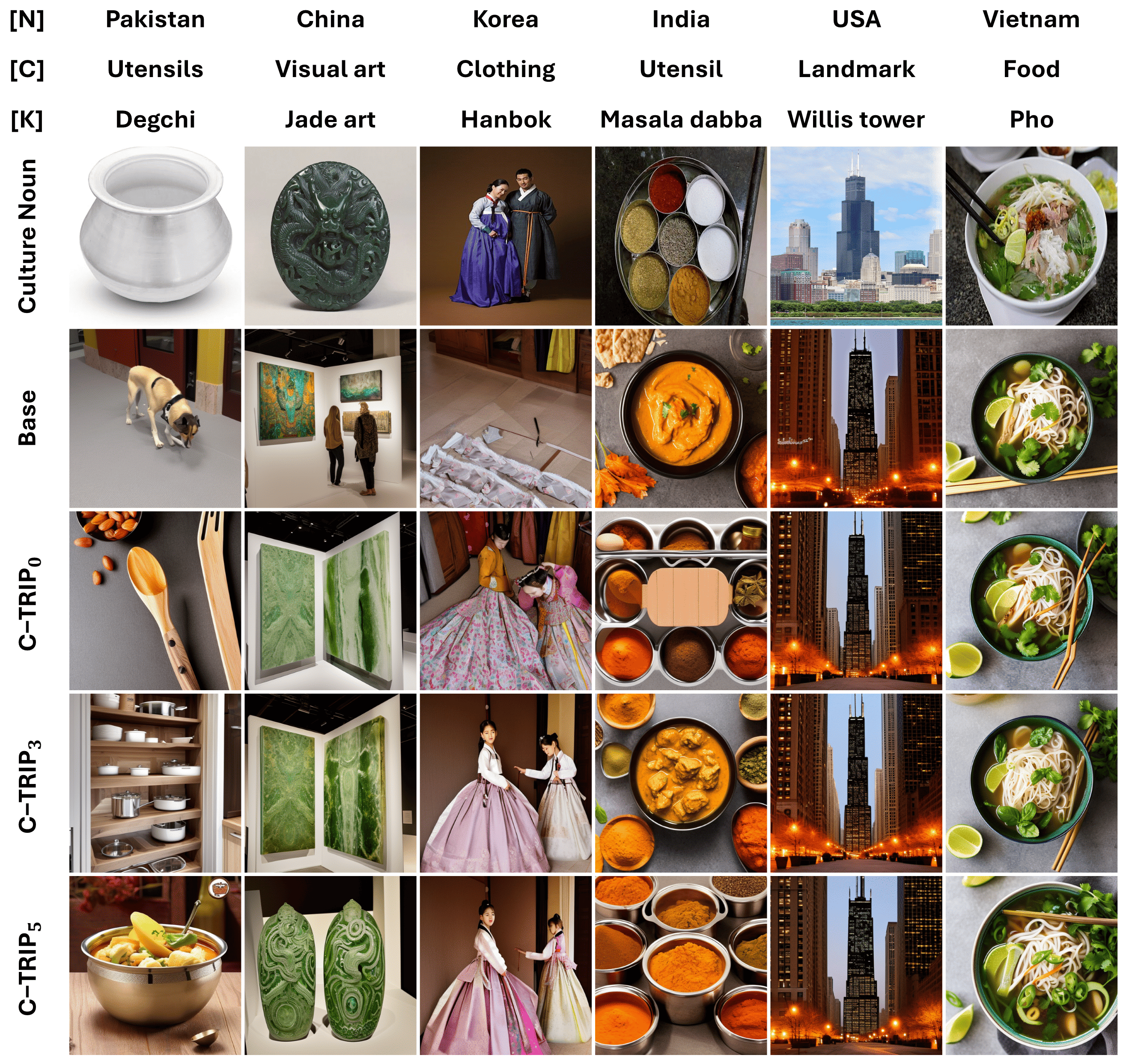}
    \caption{Qualitative comparison of C-TRIP ablated configurations compared to Base Prompt. The six columns can be divided into two groups: Relatively UC nouns (left four columns) and RC nouns (right two columns). The left group needed C-TRIP to introduce culture nouns that were underrepresented in Text-to-Image models, while the right group had to recall what they already knew through the additional information provided.}
    \label{Figure4}
\end{figure*}

\subsection{Baselines for Ablation Study}\label{sec:ablation}
To effectively analyze the specific contributions of cultural contexts and visual details based on the Culture Capsules approach, the baseline and ablation configurations were established in order to systematically examine the roles of cultural contexts (Clarity, Background, Purpose) and visual details (Visual Elements, Comparable Objects). 

In this study, we evaluated three configurations of the C-TRIP, each with different criteria for prompt refinement: C-TRIP\(_0\), C-TRIP\(_3\), and C-TRIP\(_5\). For C-TRIP\(_3\) and C-TRIP\(_5\), the refinement process was performed up to a maximum of 5 iterations, ensuring that the prompts reached the desired level of cultural and visual alignment. 

These configurations were applied to 10,000 Base Prompts, resulting in 40,000 refined prompts. Subsequently, 80,000 images were generated, with two images created for each prompt using Stable Diffusion 2 ~\citep{Rombach_2022_CVPR}.

\paragraph{C-TRIP\(_0\)} C-TRIP\(_0\) utilizes prompts augmented with raw cultural information without applying the \textit{Iterative Prompt Refinement}. This configuration is used to evaluate the baseline effect of unrefined cultural information, enabling an assessment of how much alignment can be achieved without iterative refinement.

\paragraph{C-TRIP\(_3\)} C-TRIP\(_3\)  refines the prompts based solely on the cultural context criteria (Clarity, Background, and Purpose). This setup evaluates the contribution of cultural context alone without incorporating visual details. This enables assessing how effectively the refined cultural information enhances alignment with the intended culture nouns in the generated images.

\paragraph{C-TRIP\(_5\).}
C-TRIP\(_5\) incorporates both cultural context and visual details, refining prompts according to all five criteria: Clarity, Background, Purpose, Visual Elements, and Comparable Objects. This configuration assesses whether adding visual details improves the alignment. By comparing C-TRIP\(_3\) and C-TRIP\(_5\), we can evaluate how much visual details enhance cultural alignment in the generated images.

\begin{table*}[]
\resizebox{\linewidth}{!}{%
\centering
\begin{tabular}{crcccc}
\toprule
\multicolumn{2}{c|}{Evaluation Criteria}   & Base Prompt & C-TRIP\(_0\) & C-TRIP\(_3\) & C-TRIP\(_5\) \\ \midrule
\multirow{4}{*}{User Survey (↓)} & Cultural Representation   & 2.73        & 2.53         & 2.56         & \textbf{2.18} \\
                             & The Naturalness of the Keyword  & 2.76        & 2.69         & 2.38         & \textbf{2.18} \\
                             & Offensiveness              & 2.81        & 2.60         & 2.46         & \textbf{2.13} \\
                             & Description and Image Alignment  & 2.53        & 2.66         & 2.51         & \textbf{2.30} \\ \cmidrule{1-6}
\multirow{3}{*}{VIEScore (↑)}    & Semantic Consistency (SC)            & 0.37        & 0.36         & 0.37         & \textbf{0.38} \\
                             & Perceptual Quality (PQ)             & 7.50        & \textbf{7.81}         & 7.80 & 7.76          \\
                             & Overall Score           & 0.71        & 0.71         & 0.73         & \textbf{0.74} \\ 
\bottomrule
\end{tabular}
}
\caption{Result of User Survey and VIEScore. The results highlighted in bold indicate the best outcomes. For the User Survey, a lower average rank is better, while for the VIEScore, a higher score is preferred. Except for the Perceptual Quality in the VIEScore, C-TRIP\(_5\) received the best evaluation in all other evaluations.}
\label{TableResult}
\end{table*}

\subsection{Evaluation}
\paragraph{User Survey.} The alignment of images with culture nouns is inherently subjective and can only be appropriately evaluated by participants of the respective cultural groups based on country. Accordingly, surveys were distributed to individuals who were either native to the respective countries or had at least three years of cultural experience to evaluate our approach. Participants were provided with survey questions based on their chosen country. Each survey page presented 4 images of a randomly selected culture noun. Each survey page contains 4 evaluation questions to rank images: (a) Cultural Representation, (b) The Naturalness of the Keyword, (c) Offensiveness, and (d) Description and Image Alignment. Participants were asked to evaluate a randomly ordered set of images for each question. The image ranked first was considered the most appropriately represented and least offensive, while the image ranked fourth was deemed the most inappropriate and offensive. Detailed information regarding the user survey is provided in the appendix \ref{appendix:survey}.

We employed the Matrix Mean-Subsequence Reduced (MMSR) model ~\citep{ma2020adversarial}, an established algorithm ~\citep{majdi2023crowd} for noise label aggregation provided by crowd-kit ~\citep{ustalov2021learning}, to quantitatively estimate subjective performance perception. Using MMSR, the labels from all respondents were aggregated through weighted majority voting based on the assessment of their reliability. Subsequently, the MMSR+Vote method was applied, in which labels were further aggregated and ranked using simple majority voting. 

\paragraph{Automatic Evaluation.} In addition to the cultural survey, we measured the VIEScore ~\citep{ku2023viescore} using GPT-4o, demonstrating a strong correlation with human evaluations of text-guide image generation. The VIEScore assesses the Semantic Consistency (SC) and the Perceptual Quality (PQ) and Overall score based on these metrics. To evaluate C-TRIP, 150 Base Prompts were randomly sampled from each country. These prompts were then processed through C-TRIP configurations, resulting in a total of 600 samples per country for which scores were measured.

\section{Results}
\paragraph{Qualitative Comparison}
Figure \ref{Figure4} compares the images generated from each C-TRIP configuration and the Base Prompt described in Section \ref{sec:ablation}. The refined prompt generated by C-TRIP provides cultural knowledge to Stable Diffusion 2, contributing to the generation of culturally-aware images. C-TRIP\(_5\), which includes the visual details criteria, demonstrated higher quality representation. That is, Stable Diffusion 2 can produce better images when the prompts include appropriate cultural contexts and visual details. With these enhancements, C-TRIP effectively improves the model's ability to generate culturally relevant images.


\paragraph{User Survey Results.} 
A total of 66 participants from eight countries participated in the survey, ranking the four configurations of the generated images based on the four evaluation questions specified in the survey. The average ranking for each selected configuration is presented in the Table \ref{TableResult}. 

When applying the MMSR algorithm to all the survey responses, C-TRIP\(_5\) achieved the highest ranking overall. C-TRIP\(_3\) ranked second-highest, using cultural contexts criteria alone is still effective in the refinement. However, C-TRIP\(_0\) scored lower than the Base Prompt, particularly in the Description and Image Alignment questions. This suggests that unrefined prompts may introduce irrelevant details, which can degrade description quality.

Further analyzing alignment, we converted the rankings into binary comparisons between C-TRIP\(_5\) and the Base Prompt, isolating and comparing their respective rankings. This approach revealed that C-TRIP\(_5\) demonstrated higher average alignment in 61\% of the time. 

In conclusion, incorporating both cultural contexts and, importantly, visual details in the scoring process significantly enhances the alignment of culture nouns with the generated images.

\paragraph{Automatic Evaluation Results.}
C-TRIP\(_5\) scored the best in the evaluation of the SC score, evaluating the semantic similarity between the prompt and the generated image. However, C-TRIP\(_0\) scored the best in the PQ score, which assesses the naturalness of the generated image. This result suggests that additional iterative refinement can potentially reduce the perceived naturalness of the image. It appears that the additional refinement process may have overemphasized specific details, thereby deviating from the naturalness in quality as perceived by GPT-4o. Nevertheless, C-TRIP\(_5\) still outperformed the Base Prompt by 3.4\%. Overall, C-TRIP\(_5\) maintained the highest score. In conclusion, as in the user survey results, incorporating both cultural and visual refinements consistently enhances the overall performance.

\section{Ablation Study for UC Nouns}
The Culture Capsule approach is a method that teaches learners who have not experienced a particular culture. With this idea in mind, we aimed to apply a similar approach to words that Stable Diffusion 2 is not familiar with. In this section, we analyzed and compared the Unrecognized/Underrepresented Culture nouns (UC nouns) and the Common/Recognized Culture nouns (RC nouns).

\begin{figure}[t]
    \centering
    \includegraphics[width=0.45\textwidth]{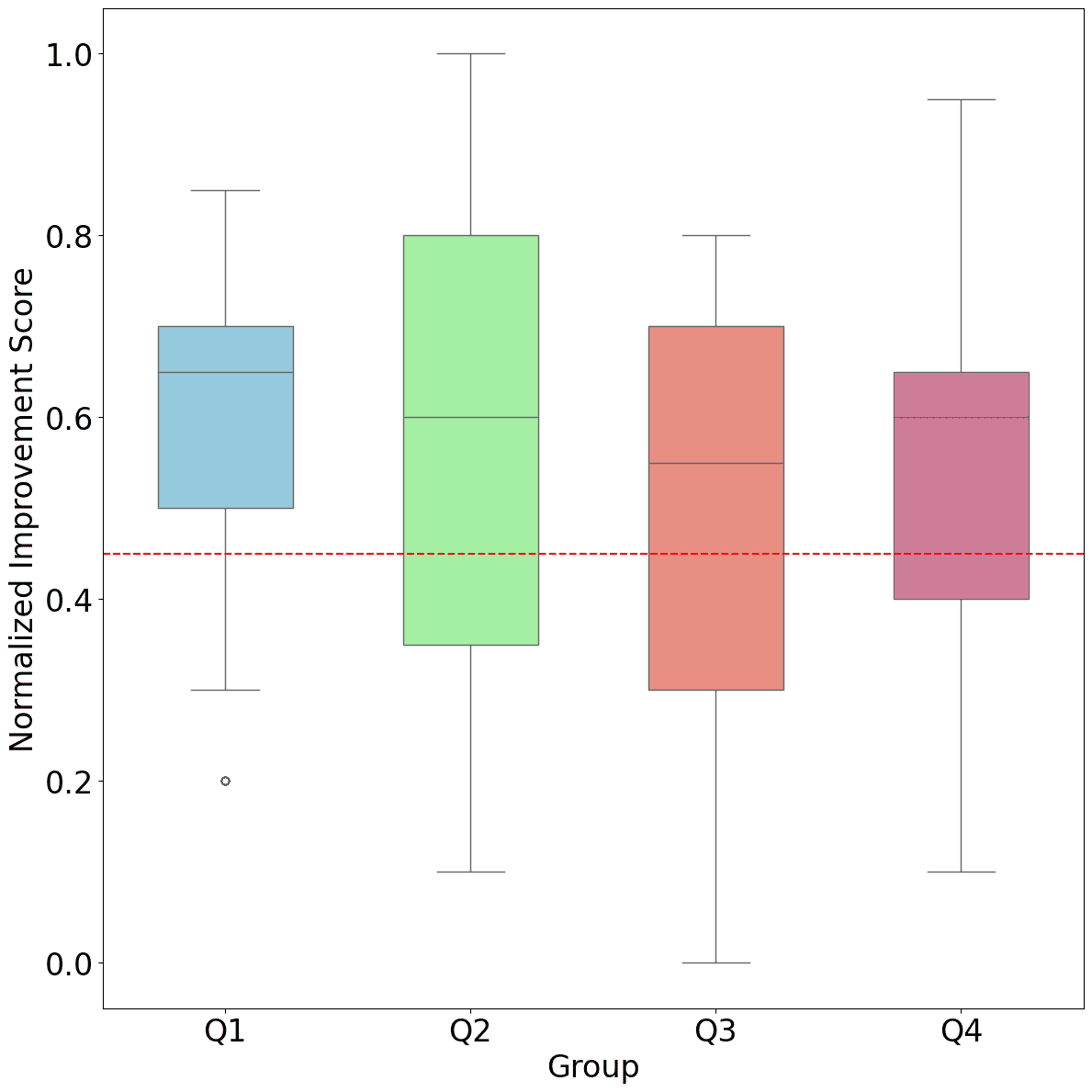}
    \caption{A box plot illustrating the normalized improvement scores for each group (Q1, Q2, Q3, and Q4). A score exceeding 0.45 signifies that the C-TRIP's guidelines enhance the image alignment of the Stable Diffusion 2 model. Notably, the Q1 group exhibits the highest performance improvement compared to the other groups.}
    \label{Relatve Score}
\end{figure}

\paragraph{UC and RC noun groups}
We categorized culture nouns into UC and RC noun groups according to their frequency within the training dataset used for Stable Diffusion 2. To achieve this, we analyzed Re-LAION-2B-en-research, a filtered true subset of the LAION-2B-en \citep{schuhmann2022laion}, using a \(p_{unsafe}>0.95\) threshold and keyword-based filters to remove potentially suspicious content.\footnote{\url{https://laion.ai/blog/relaion-5b/}}

Our analysis focused on culture nouns within the dataset captions, classifying them based on their frequency of appearance. These nouns were then grouped using a quartile-based approach (Q1 to Q4), with Q1 and Q2 representing UC nouns and Q3 and Q4 representing RC nouns. This grouping provided a structured means to evaluate the C-TRIP’s capacity to generate culturally aligned images across varying levels of representation.

\paragraph{User Survey.}
We used the normalized improvement score to evaluate UC nouns. This score is calculated by normalizing the difference between the average rankings of C-TRIP\(_5\) and the Base Prompt in the user survey, measuring the improvement of C-TRIP\(_5\) over the Base Prompt. A normalized improvement score higher than 0.45 indicates that C-TRIP's guidelines improved image alignment in Stable Diffusion 2.

Our approach performed the best with the Q1 group, with the highest median score and a narrow IQR, presenting consistent improvement across this group, as shown in Figure \ref{Relatve Score}. This suggests that C-TRIP effectively reinforces the generation of images for the Q1 group, enhancing alignment for UC nouns. The Q2 group demonstrated the highest upper range within the IQR and the second-highest median, suggesting effective but slightly variable improvements. However, the Q3 group showed the lowest performance with a wide variance, which may potentially indicate that additional information could decrease alignment for culture nouns.

Our approach demonstrated an 11.05\% higher mean normalized improvement score for UC nouns compared to RC nouns. The t-test yielded a t-statistic of 2.951 and a p-value of 0.0033, providing statistical evidence of a significant performance improvement for the UC noun group. These results indicate that C-TRIP's guidance was more effective in improving alignment with UC nouns than with RC nouns.

\begin{table}[t]
\resizebox{\linewidth}{!}{%
\begin{tabular}{ccccc}
\hline
                     & Q1              & Q2      & Q3      & Q4     \\ \hline \hline
Semantic Consistency (SC) & \textbf{0.1304} & -0.0003 & -0.0077 & 0.047  \\
Perceptual Quality (PQ)   & \textbf{0.4517} & 0.2879  & 0.3059  & 0.1093 \\
Overall        & \textbf{0.2158} & 0.0165  & 0.0489  & 0.0541 \\ \hline
\end{tabular}
}
\caption{Score Differences between C-TRIP\(_5\) and Base Prompt across Groups (Q1, Q2, Q3, and Q4), Assessed by GPT-4o. The VIEScore incorporates three key aspects: Semantic Consistency, reflecting the alignment between the image and the prompt; Perceptual Quality, measuring the naturalness of the generated image; and the Overall Score, which combines these metrics.}
\label{Q_VIE}
\end{table}


\paragraph{Automatic Evaluation.}
As shown in Table \ref{Q_VIE}, C-TRIP\(_5\) demonstrated improvements in all scores: SC, PQ, and Overall score. The table highlights significant gains in Q1, with notable increases across all scores. However, C-TRIP\(_5\) scored lower in the SC score with Q2 and Q3. This discrepancy with the user survey highlights the potential limitations of using VIEScore, which relies on LLMs trained on culturally biased web-based data. It emphasizes the need for surveys conducted by members of the respective cultural groups when evaluating culture nouns. The consistently low SC scores suggest that LLMs may struggle to accurately assist in recognizing and aligning culture nouns, further limiting VIEScore's effectiveness in assessing culturally specific content.

\section{Conclusion}
In this paper, we introduced\textbf{ C-TRIP (Culturally-Aware Text-to-Image Generation with Iterative Prompt Refinement)}, a novel approach that iteratively refines prompts to improve the alignment of culture nouns with images generated by existing text-to-image models without any fine-tuning. 
Experiments across eight countries demonstrated that C-TRIP significantly improves the alignment of culture nouns in generated images, particularly for underrepresented UC nouns. User surveys and automatic evaluations consistently present C-TRIP's superior performance in cultural representation and the semantic consistency.

\section{Limitations}
Sources like Wikipedia and general Web content contain cultural biases \citep{miquel2018wikipedia, baeza2018bias}, which can affect the refinement process and C-TRIP's capacity to provide balanced cultural representation. Future work should focus on enhancing the information retrieval process through developing culturally diverse datasets, thereby ensuring high-quality, relevant data for effective prompt refinement.

The limited scope of prompts utilized in our experiments and human evaluations presents a current limitation and suggests an important avenue for future research. Additionally, our work is constrained by the perceptual biases of human annotators from eight countries. To improve the reliability of evaluation outcomes, future work will emphasize the inclusion of annotators from a broader range of cultural backgrounds.

\paragraph{Acknowledgements}
This research was supported by the MSIT(Ministry of Science and ICT), Korea, under the ITRC(Information Technology Research Center) support program(IITP-2025-2020-0-01789), the Artificial Intelligence Convergence Innovation Human Resources Development (IITP-2025-RS-2023-00254592) supervised by the IITP(Institute for Information \& Communications Technology Planning \& Evaluation), and the Hyundai Motor Chung Mong-Koo Foundation. 


\bibliography{custom}

\appendix
\section{List of Culture Nouns}
This part lists culture nouns for the eight countries used in the experiment. The culture nouns are divided into 8 categories: architecture (3), city \& landmark (5), clothing (4), dance \& music (2), visual arts (1), food \& drink (5), religion \& festival (3), and utensils \& tools (2). The numbers in parentheses indicate the quantity of nouns used in each category. A total of 25 nouns were extracted and used for each country as shown in Table \ref{tab5} to \ref{tab12}.

\section{Retrieve Cultural Information}
This part covers the process of retrieving cultural information. As mentioned in the text, the retrieval is carried out in two steps. First, information about the culture noun is retrieved from Wikipedia. However, for culture nouns categorized under UC nouns, when information is unavailable or insufficient on Wikipedia, further information is retrieved from the web. Both retrieval processes utilized LangChain modules, with WikipediaQueryRun\footnote{\url{https://python.langchain.com/api_reference/community/tools/langchain_community.tools.wikipedia.tool.WikipediaQueryRun.html}} used for Wikipedia searches and GoogleSearchAPIWrapper\footnote{\url{https://python.langchain.com/docs/integrations/tools/google_search/}} used for web searches. The two examples below show cases where the retrieval was successful through Wikipedia as shown in Figure \ref{Figure9} and where the information was not insufficient through Wikipedia, so additional retrieval was also conducted from the web as shown in \ref{Figure10}.

\section{Iterative Prompt Refinement}\label{sec:Iterative Prompt Refinement}
Figure \ref{Figure11} visually depicts how the prompt evolves through the Iterative Prompt Refinement process. First, the prompt is refined, and the score is measured based on the five evaluation criteria defined in Table \ref{Table1}. Next, the feedback process generates an explanation for the given score, based on the refined prompt and the measured score. Finally, the prompt is iteratively refined and scored based on the feedback. The Iterative Prompt Refinement process terminates when the total score of the refined prompt exceeds 40 points or after 5 iterations. The Refine, Score, and Feedback processes all utilize LLaMA-3-70B, and the respective templates can be found in Figure \ref{Template1}, Figure \ref{Template2}, and Figure \ref{Template3}.

\section{Qualitative Comparison Prompts}
This part includes the prompts used to generate the images in Figure \ref{Figure4} of the main text. \textcolor{orange}{Cultural contexts} are written in \textcolor{orange}{orange}, and \textcolor{cyan}{visual details} are written in \textcolor{cyan}{cyan.} For culture nouns categorized under the UC nouns (dengchi, jade art, and masala dabba), we observe that Stable Diffusion 2 effectively aligns with the culture nouns using prompt refined through our approach. 

In contrast, for culture nouns categorized under the RC nouns (Willis Tower and pho), there is no significant difference in the images generated between the refined prompt and the base prompt. For pho (Vietnamese cuisine), Stable Diffusion 2 generates images similar to the original reference image, even without visual details in the refined prompt.
\section{User Survey} \label{appendix:survey}
 To evaluate the performance of our approach, we recruited 66 participants with at least 3 years or older of cultural experience in each of the 8 countries. Even so, to address the mitigate bias further, we recruited at least 5 participants from each country. Table \ref{Table3} provides detailed information about the participants. Each participant responded to 15 survey pages. A single page of the survey form includes culture nouns, a base prompt, and one image generated from each of the three approaches described in Section \ref{sec:ablation}, for a total of four images. Each survey page has a total of four survey items (see Table \ref{survey_item}) to rank relative to (a) Cultural Representation, (b) The naturalness of the keyword, (c) Offensiveness, and (d) Description and Image Alignment. A sample of a survey page can be viewed in Figure \ref{Figure12}. Survey participants were compensated for their time and contributions in accordance with ethical guidelines.

\section{Additional Qualitative Samples}
In Figure \ref{UC_RC1} to \ref{UC_RC8}, we present additional qualitative examples illustrating how the C\_TRIP approach improves the alignment of culture nouns in UC and RC nouns with the generated images by Stable Diffusion 2. Results are showcased across China, Germany, India, Japan, Pakistan, South Korea, USA, and Vietnam. Additionally, our approach effectively enhanced the alignment between prompt containing culture nouns from the UC nouns and the generated images.

\section{Culture noun Distribution by Quartile Group for Each Country in the Re-LAION Dataset}
Table \ref{Table14} presents the culture noun counts for each country, divided into quartiles. Each quartile represents 25\% of the data, helping to better understand and compare the frequency of culture nouns across countries. This allows for an analysis of the characteristics of cultural expression distribution by country, making it easier to identify the proportion of UC and RC nouns.

\begin{figure} [b]
    \centering
    \includegraphics[width=0.4\textwidth]{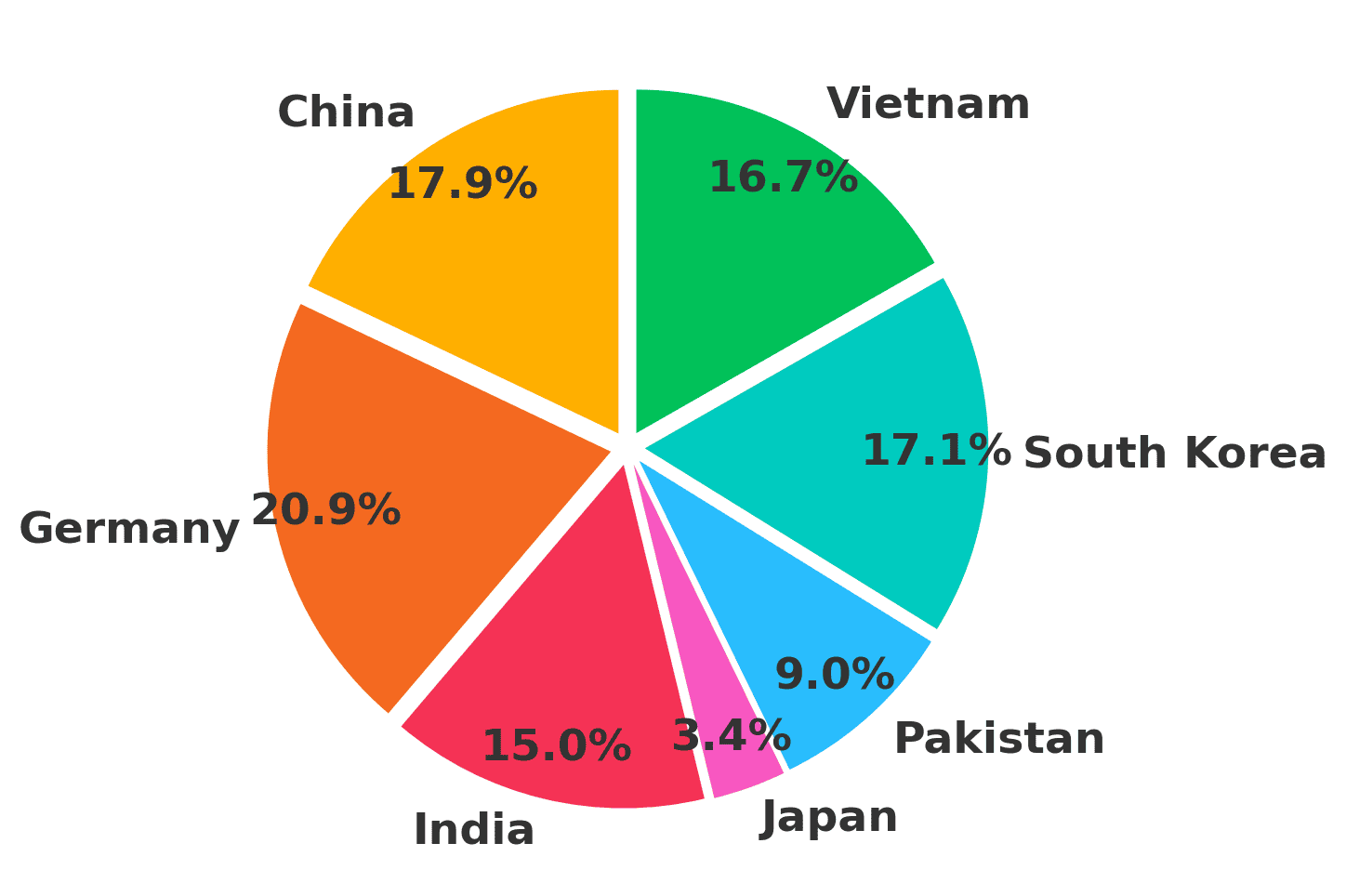}
    \caption{Country Distribution in Q1 group.}
    \label{Q1}
\end{figure}

\begin{figure}
    \centering
    \includegraphics[width=0.4\textwidth]{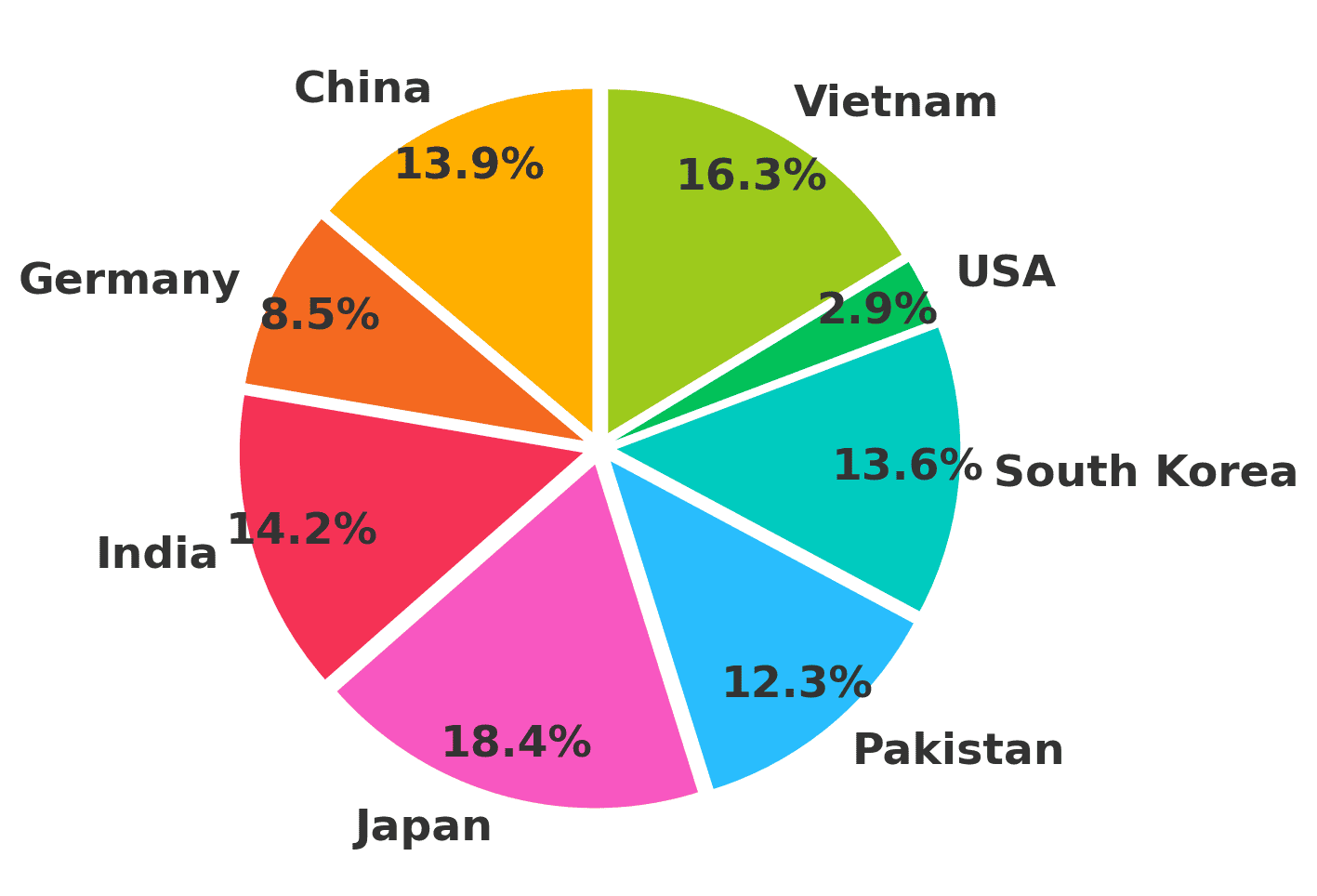}
    \caption{Country Distribution in Q2 group.}
    \label{Q2}
\end{figure}

\begin{figure}
    \centering
    \includegraphics[width=0.4\textwidth]{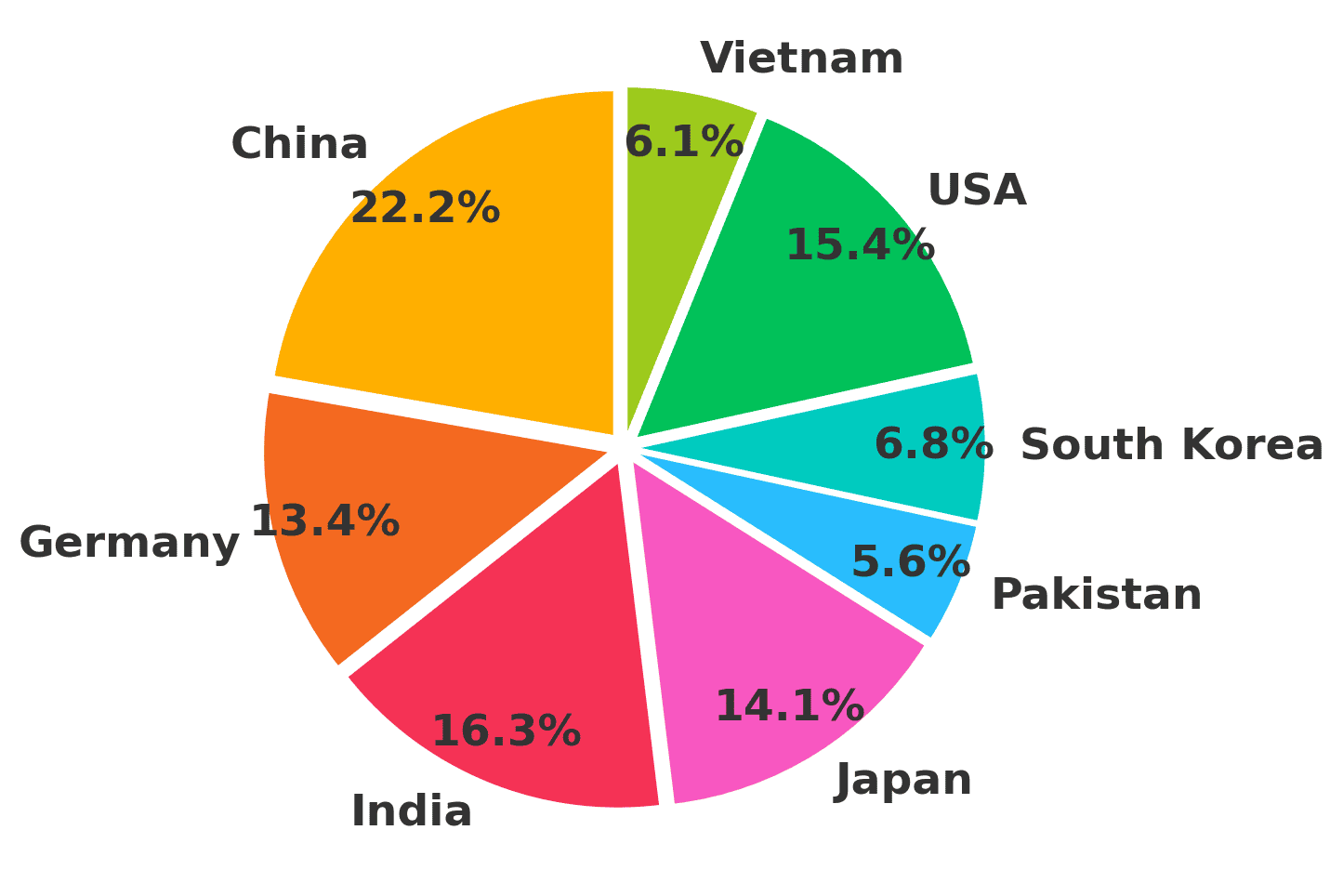}
    \caption{Country Distribution in Q3 group.}
    \label{Q3}
\end{figure}

\begin{figure}
    \centering
    \includegraphics[width=0.3\textwidth]{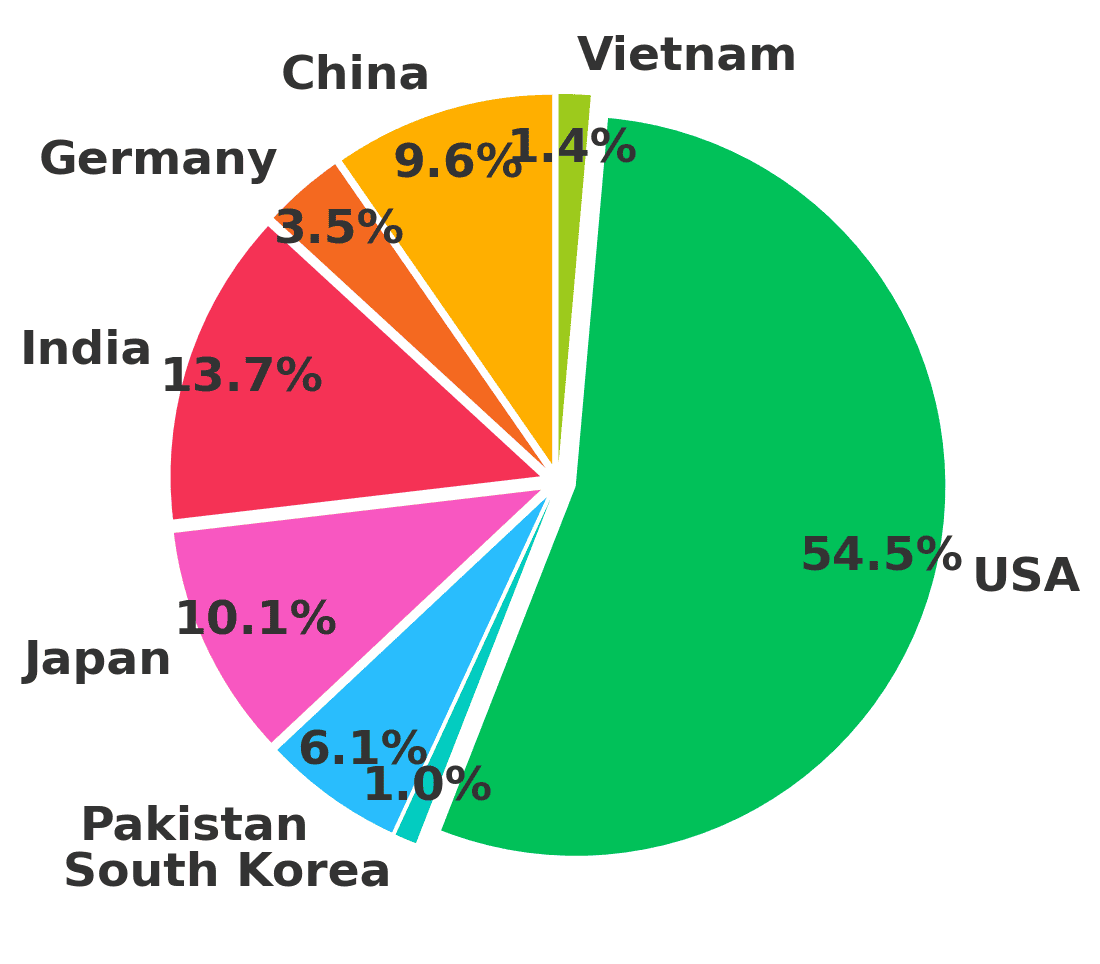}
    \caption{Country Distribution in Q4 group.}
    \label{Q4}
\end{figure}

\begin{figure} [H]
    \centering
    \includegraphics[width=0.75\linewidth]{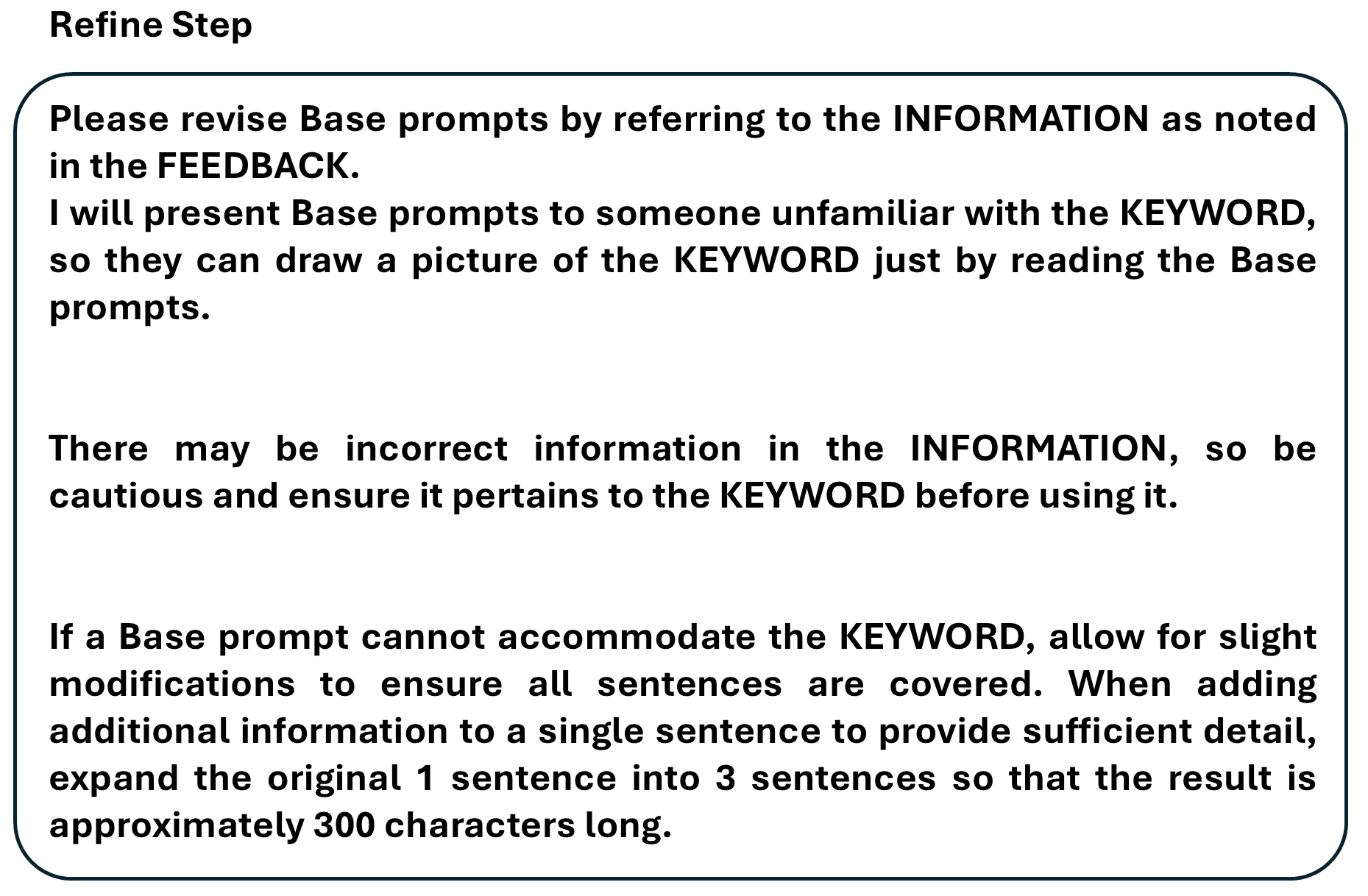}
    \caption{Prompt provided to LLaMA-3-70B in the Refine step. There are two key points in the Refine step. The first is to effectively refine the INFORMATION and incorporate it into the Base prompt. To address the first key point, we structured the first and second paragraphs. The second is to inject the information while maintaining the structure of the Base prompt. At this stage, we also imposed a length limit to prevent exceeding Stable Diffusion's input limit and to ensure that the Base prompt remains the focal point, avoiding distraction from too much information. For the second key point, we structured the third paragraph.}
    \label{Template1}
\end{figure}

\begin{figure} [H]
    \centering
    \includegraphics[width=0.75\linewidth]{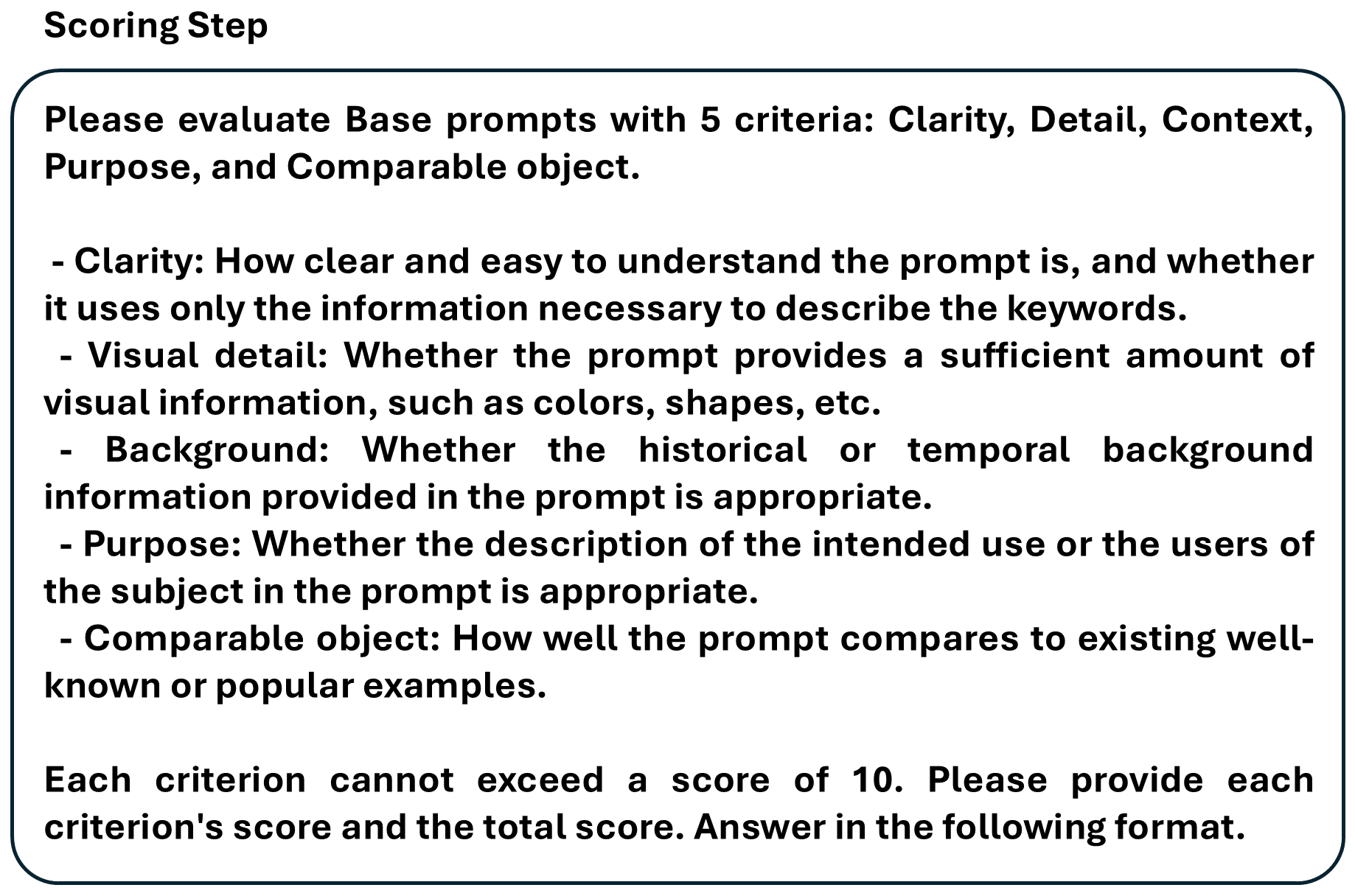}
    \caption{Prompt provided to LLaMA-3-70B in the Scoring step. We provided LLaMA-3-70B with the criteria and definitions set in Table \ref{Table1} to score each criterion up to 10 points, for a total maximum score of 50 points.}
    \label{Template2}
\end{figure}

\begin{figure} [H]
    \centering
    \includegraphics[width=0.75\linewidth]{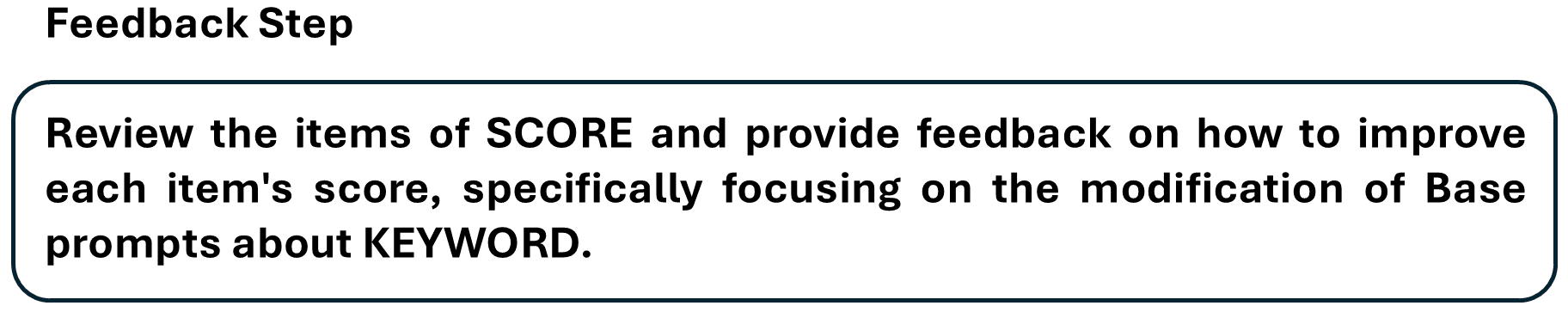}
    \caption{Prompt provided to LLaMA-3-70B in the Feedback step. We focused on reviewing the scores from the Scoring step and generating feedback to improve areas with lower scores.}
    \label{Template3}
\end{figure}

\begin{figure} [H]
    \centering
    \includegraphics[width=1\textwidth]{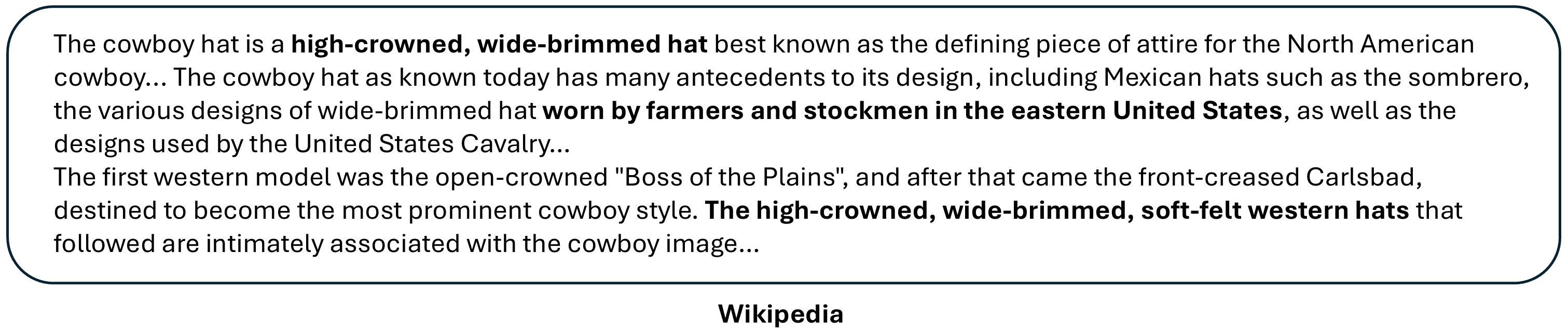}
    \caption{Example of sufficient information retrieved from Wikipedia: Cowboy hat (USA clothes). This culture noun, classified within the Q4 group and categorized under RC nouns, provided detailed information on its appearance and cultural significance, accessible solely through Wikipedia.}
    \label{Figure9}
\end{figure}

\begin{figure} [H]
    \centering
    \includegraphics[width=1\textwidth]{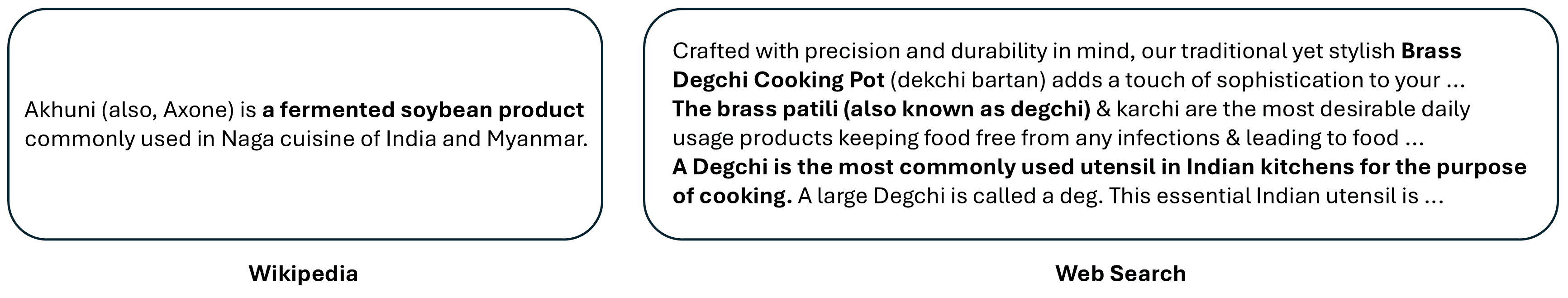}
    \caption{Example of insufficient information retrieved from Wikipedia: Degchi (Pakistan utensil). This culture noun, classified within the Q1 group and categorized under UC nouns, could not be found on Wikipedia. Additional cultural information was subsequently obtained through a web search.}
    \label{Figure10}
\end{figure}

\begin{figure} [H]
    \centering
    \includegraphics[width=1\textwidth]{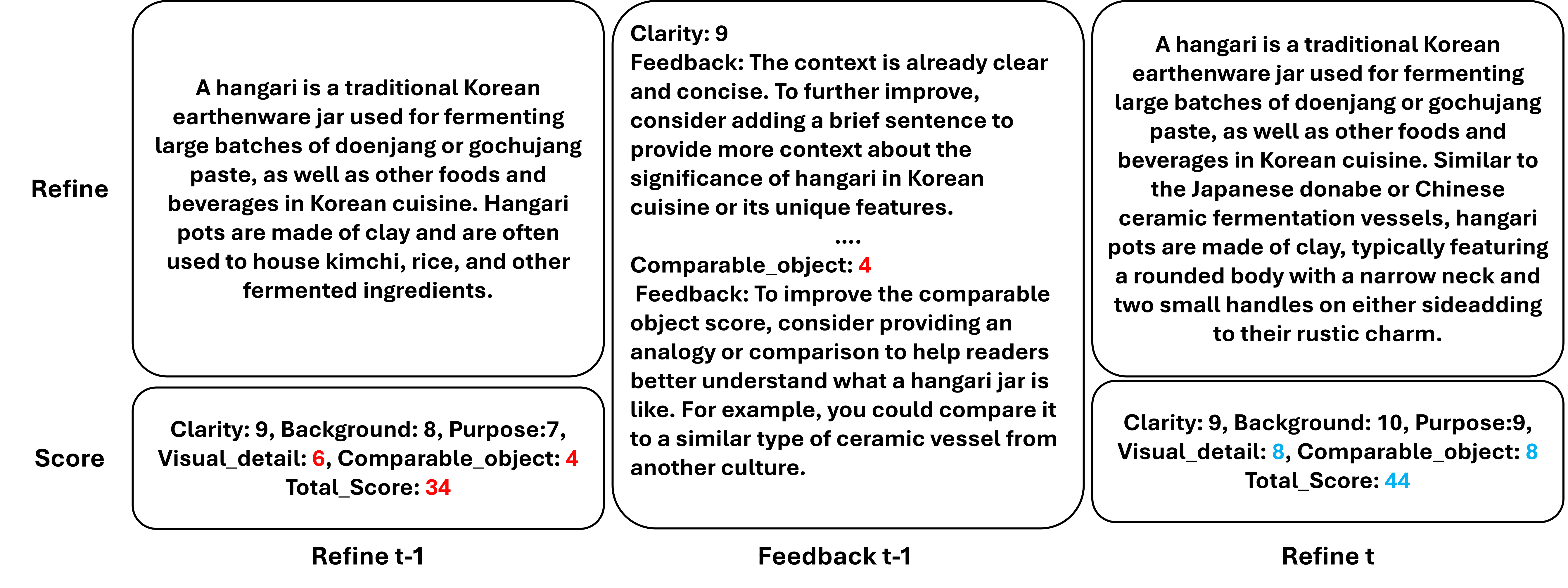}
    \caption{The step-by-step results of C-TRIP. The example uses "hangari", a Korean conventional utensil. In the (t-1)th iteration, a score is assigned based on the criteria, and feedback is generated based on this score. Then, in the t-th iteration, the prompt is revised using the feedback.}
    \label{Figure11}
\end{figure}

\begin{figure} [H]
    \centering
    \includegraphics[width=1\textwidth]{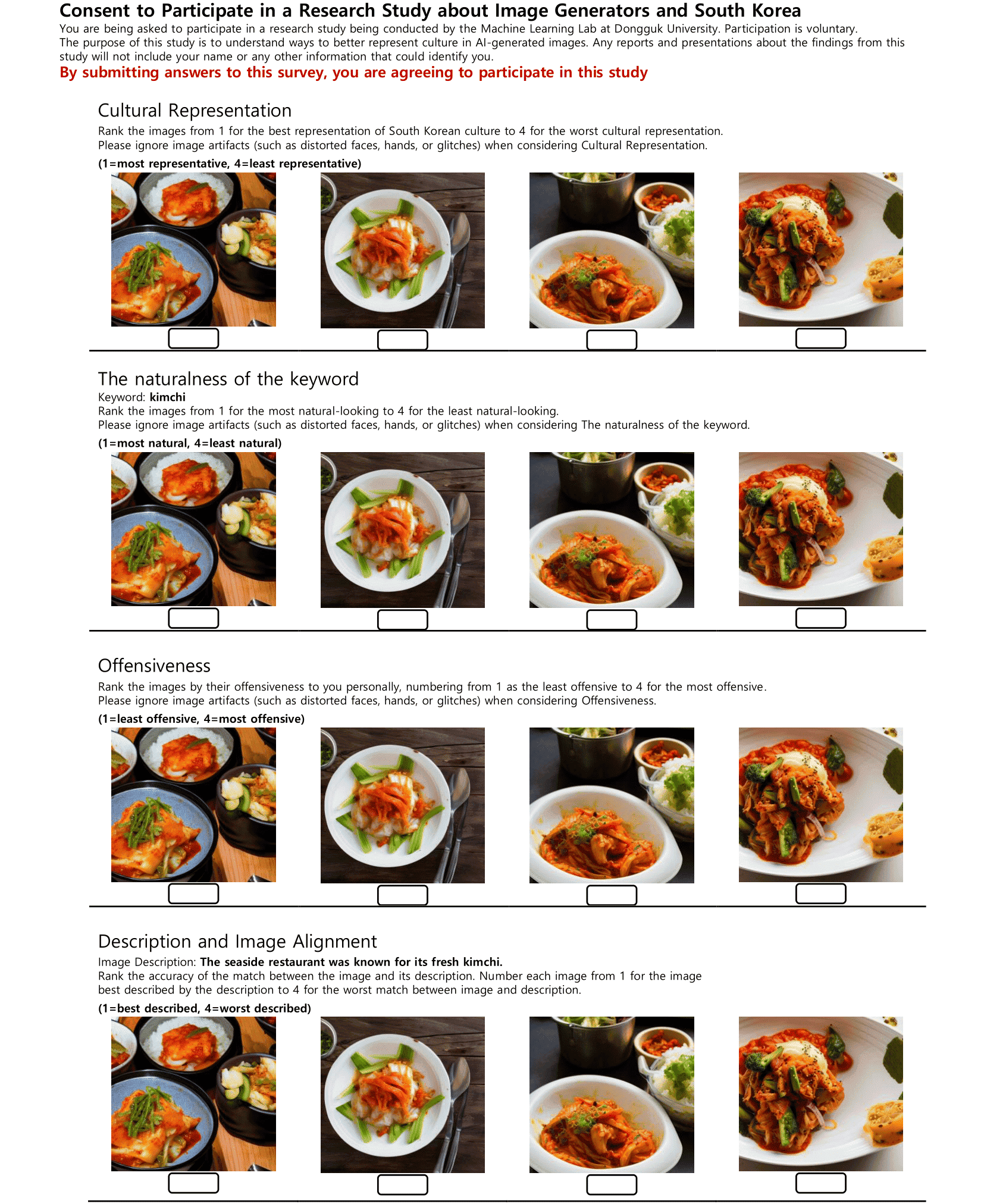}
    \caption{A sample page from the survey presents participants with four images, and below, they are asked to enter a rank between 1 and 4 into designated text fields. Participants provide evaluations based on four criteria: Cultural Representation, Keyword Naturalness, Offensiveness, and Alignment between Description and Image. For each survey item, four images are displayed in a randomized but consistent order throughout the page.}
    \label{Figure12}
\end{figure}

\begin{figure}
    \centering
    \includegraphics[width=\textwidth]{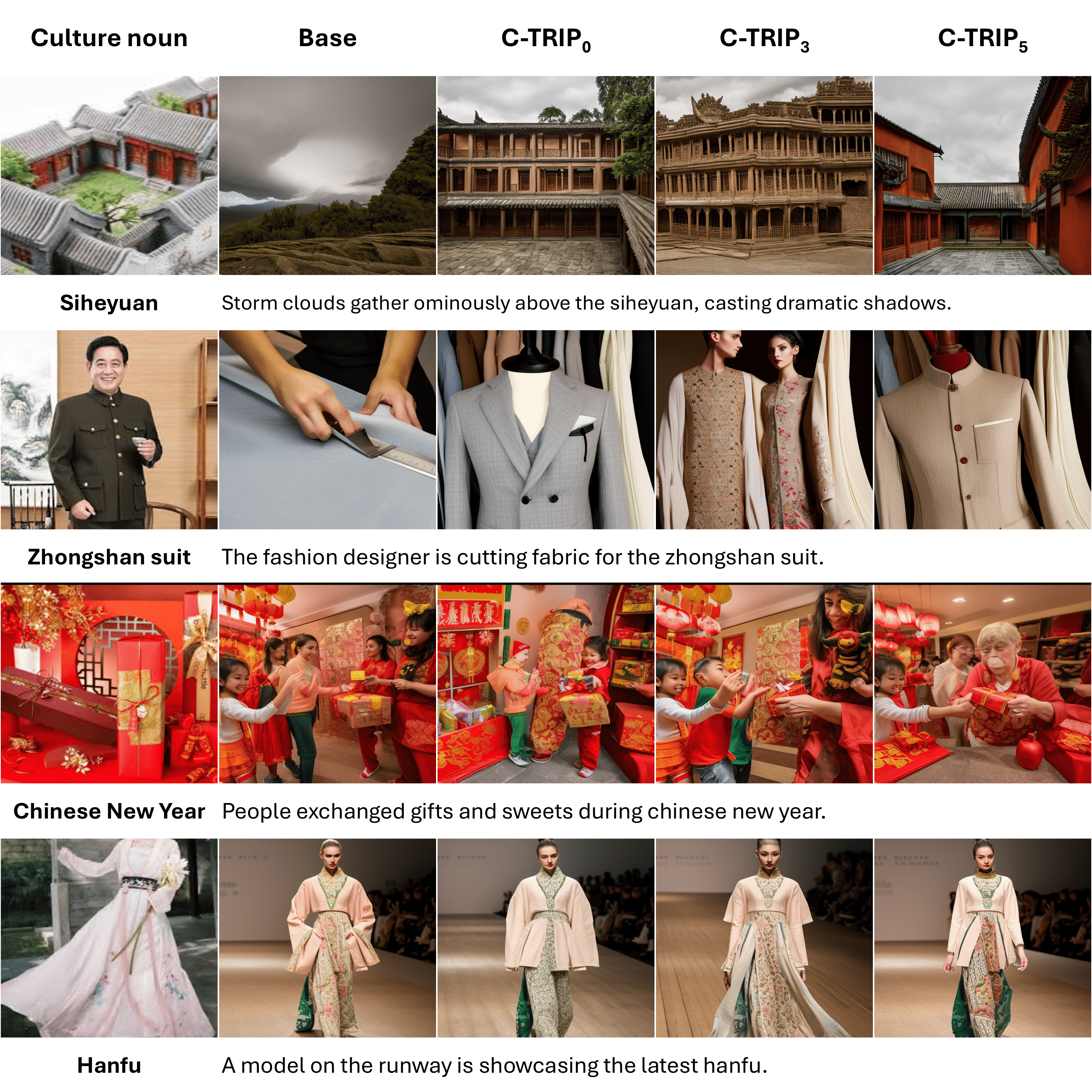}
    \caption{\textbf{Additional Qualitative Sample for Chinese culture.} The top two images show the results of C-TRIP for culture nouns categorized under UC nouns, while the bottom two images present the result for culture nouns categorized under RC nouns.}
    \label{UC_RC1}
\end{figure}

\begin{figure} 
    \centering
    \includegraphics[width=\textwidth]{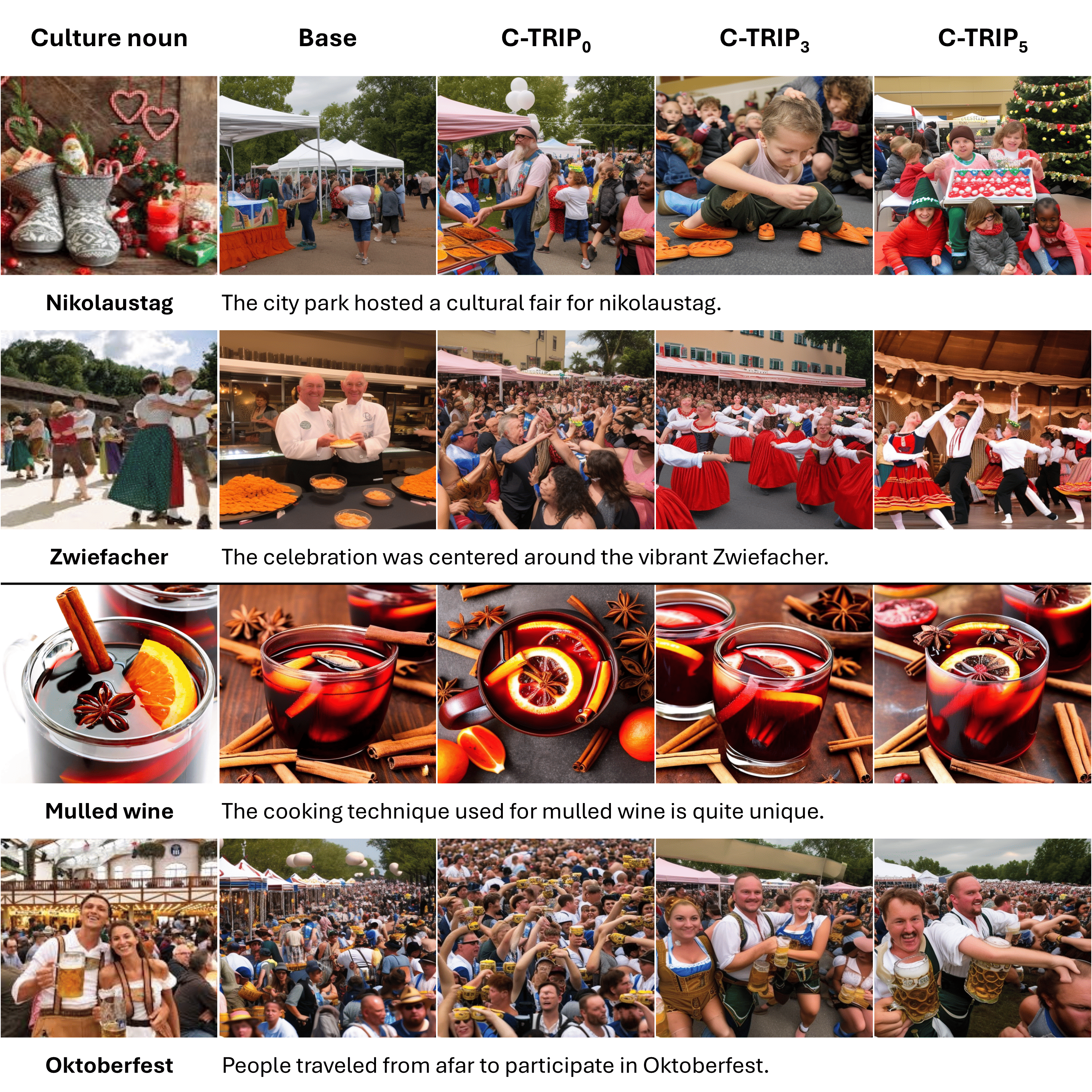}
    \caption{\textbf{Additional Qualitative Sample for German culture.} The top two images show the results of C-TRIP for culture nouns categorized under UC nouns, while the bottom two images present the result for culture nouns categorized under RC nouns.}
    \label{UC_RC2}
\end{figure}

\begin{figure}  
    \centering
    \includegraphics[width=\textwidth]{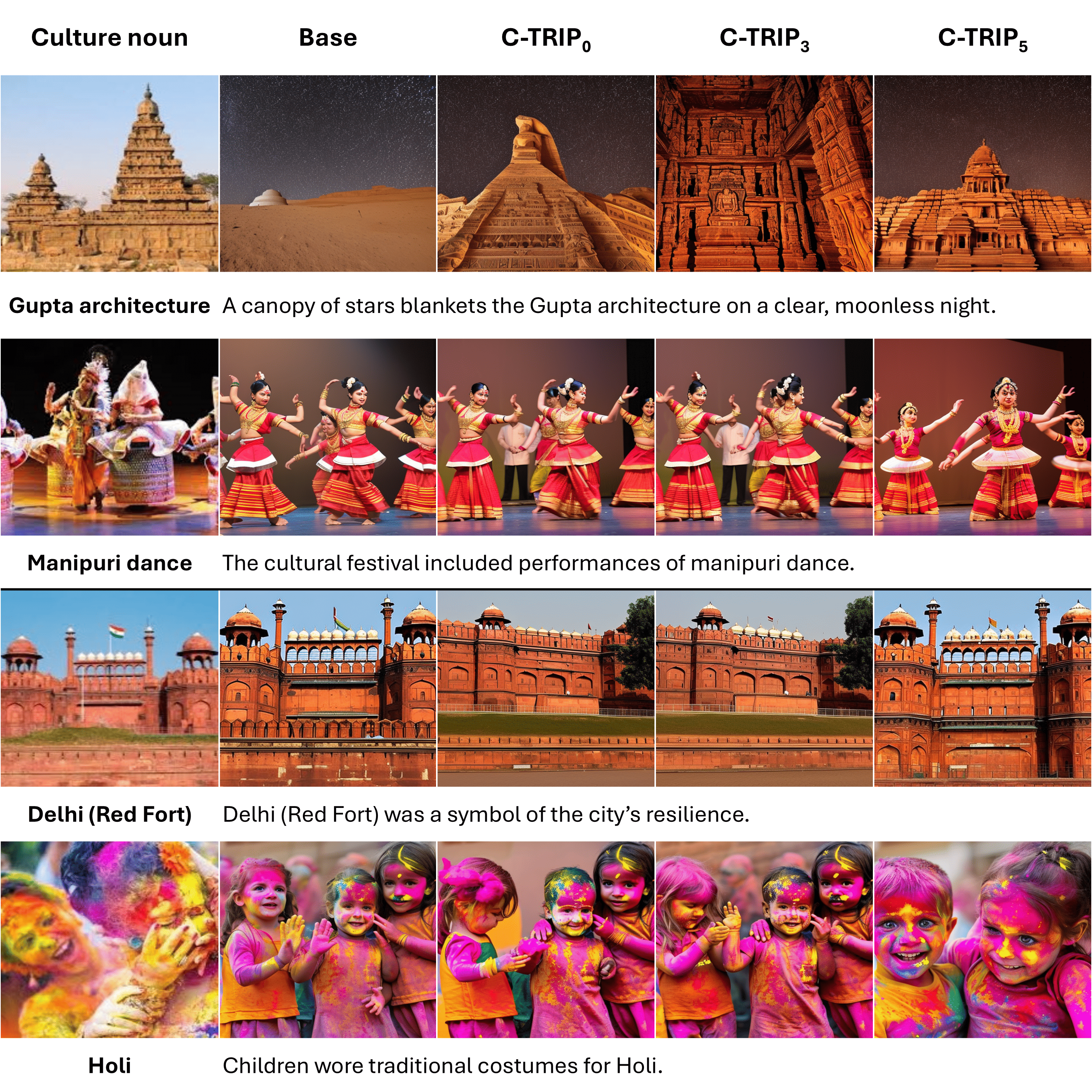}
    \caption{\textbf{Additional Qualitative Sample for Indian culture.} The top two images show the results of C-TRIP for culture nouns categorized under UC nouns, while the bottom two images present the result for culture nouns categorized under RC nouns.}
    \label{UC_RC3}
\end{figure}

\begin{figure}  
    \centering
    \includegraphics[width=\textwidth]{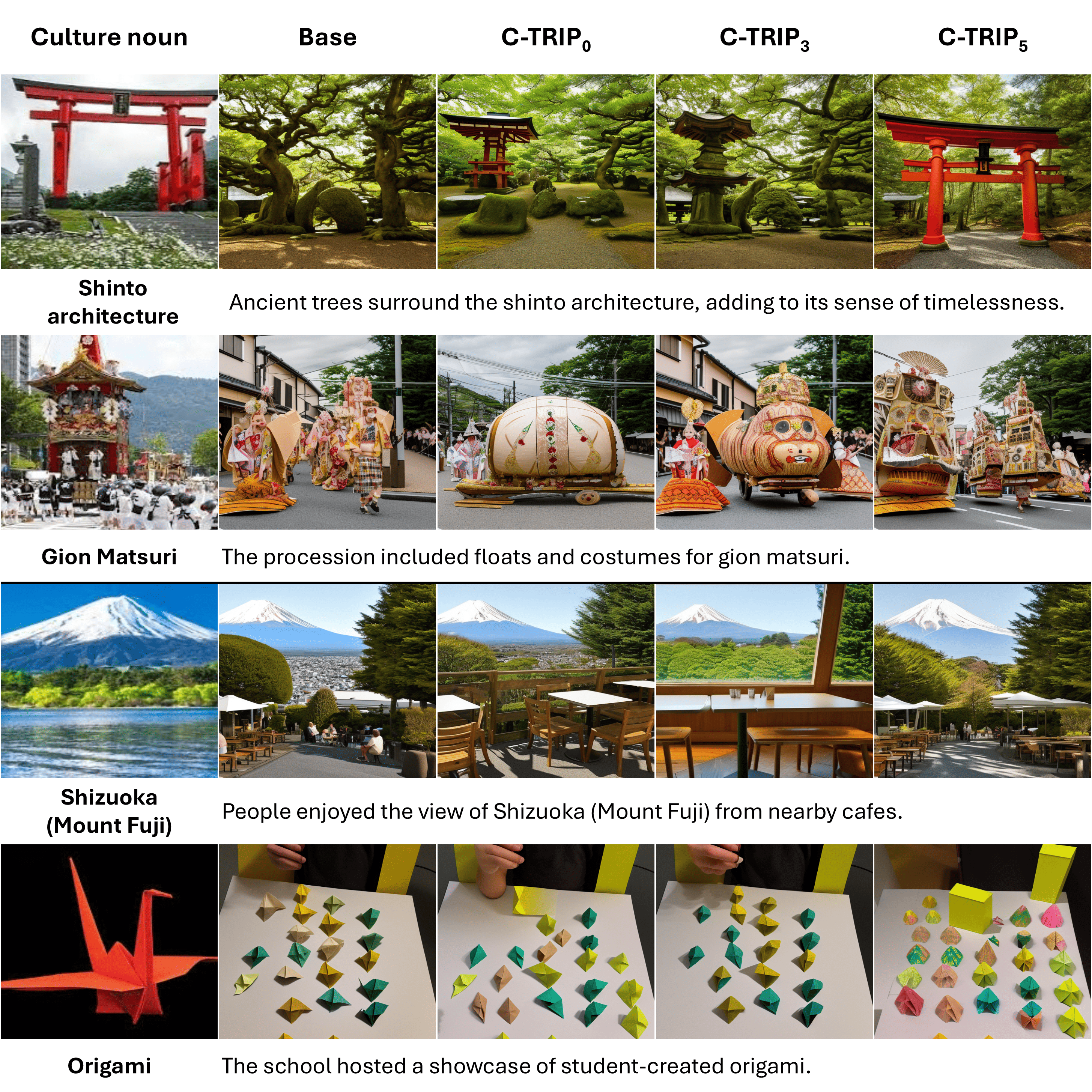}
    \caption{\textbf{Additional Qualitative Sample for Japanese culture.} The top two images show the results of C-TRIP for culture nouns categorized under UC nouns, while the bottom two images present the result for culture nouns categorized under RC nouns.}
    \label{UC_RC4}
\end{figure}

\begin{figure}  
    \centering
    \includegraphics[width=\textwidth]{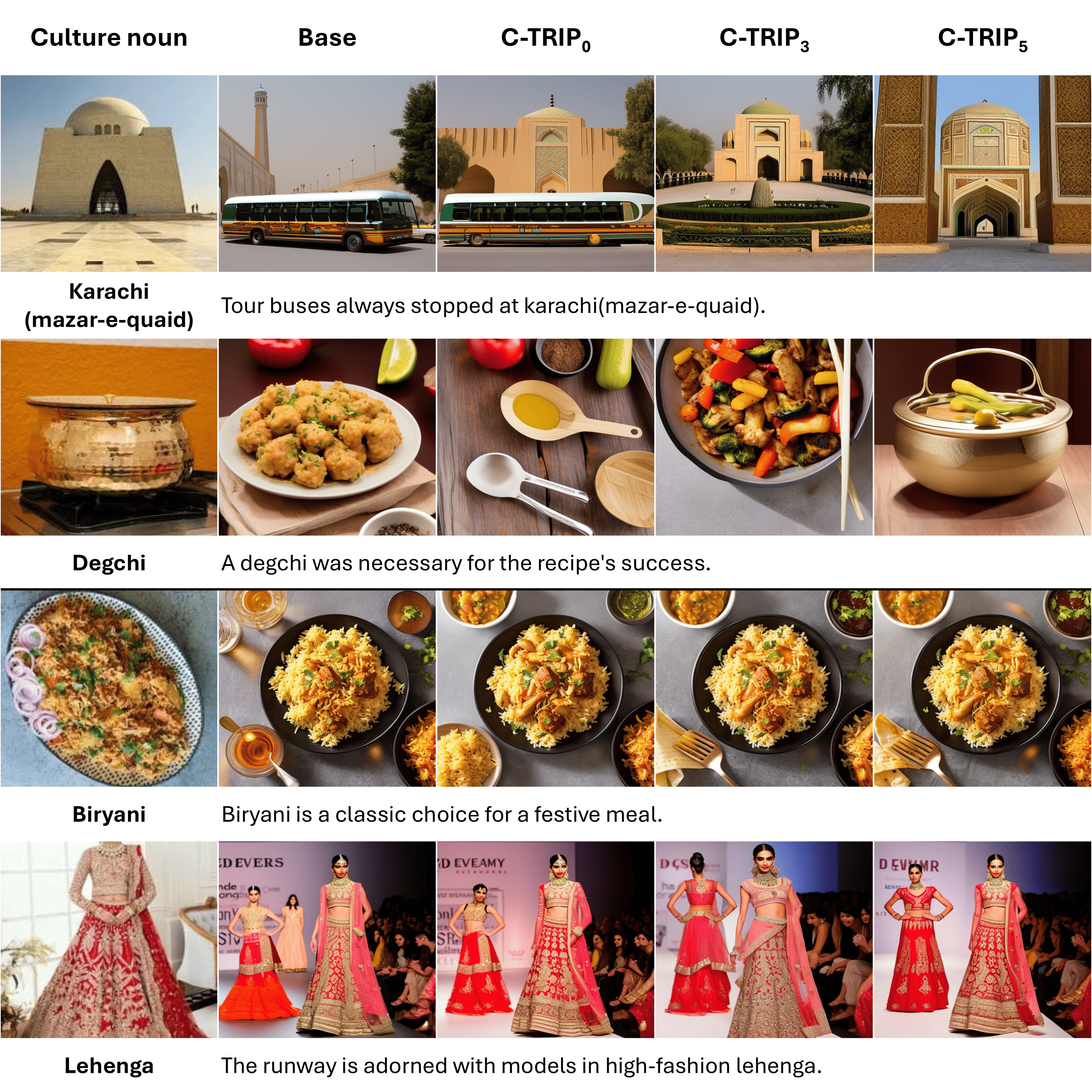}
    \caption{\textbf{Additional Qualitative Sample for Pakistani culture.} The top two images show the results of C\_TRIP for culture nouns categorized under UC nouns, while the bottom two images present the result for culture nouns categorized under RC nouns.}
    \label{UC_RC5}
\end{figure}

\begin{figure}
    \centering
    \includegraphics[width=\textwidth]{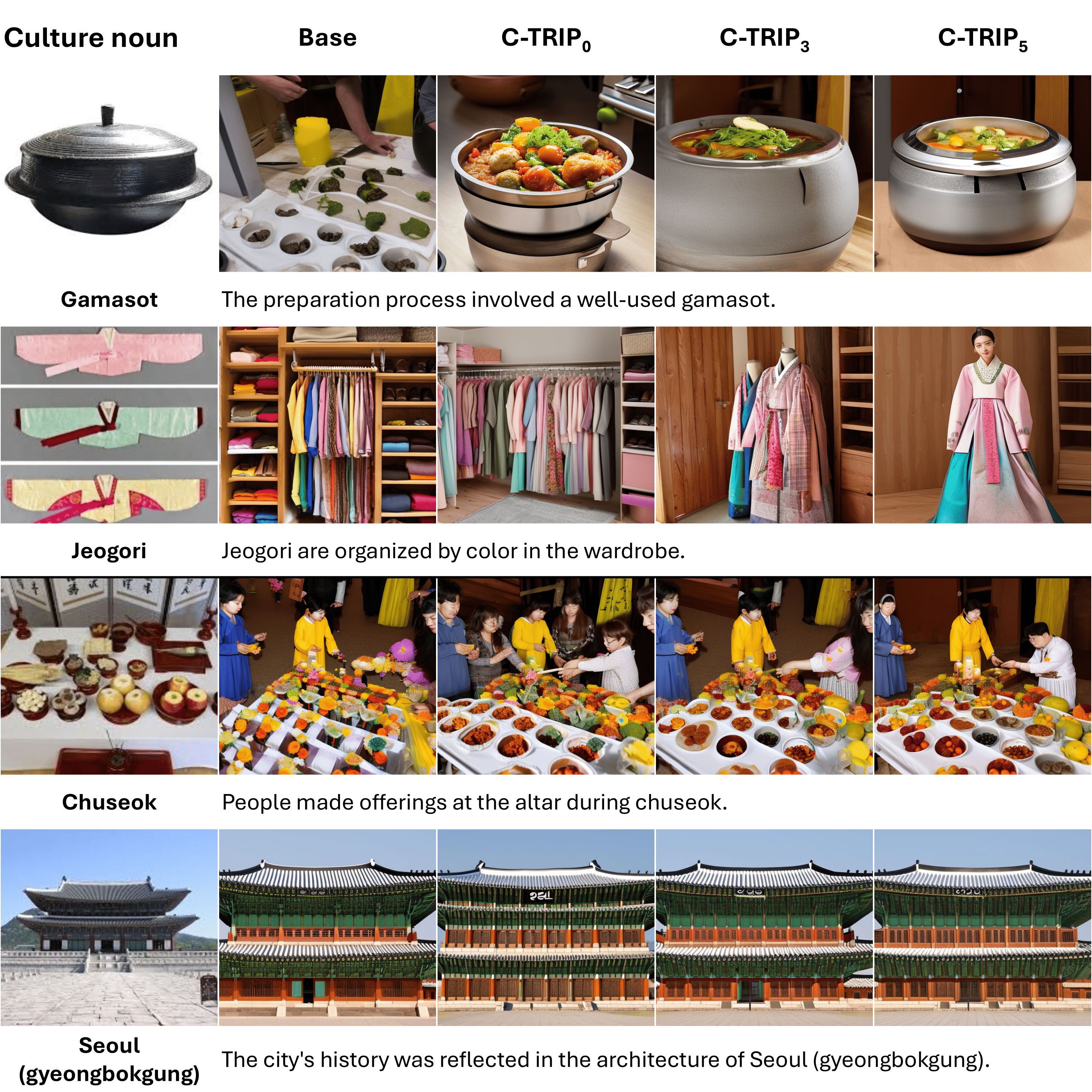}
    \caption{\textbf{Additional Qualitative Sample for Korean culture.} The top two images show the results of C-TRIP for culture nouns categorized under UC nouns, while the bottom two images present the result for culture nouns categorized under RC nouns.}
    \label{UC_RC6}
\end{figure}

\begin{figure} 
    \centering
    \includegraphics[width=\textwidth]{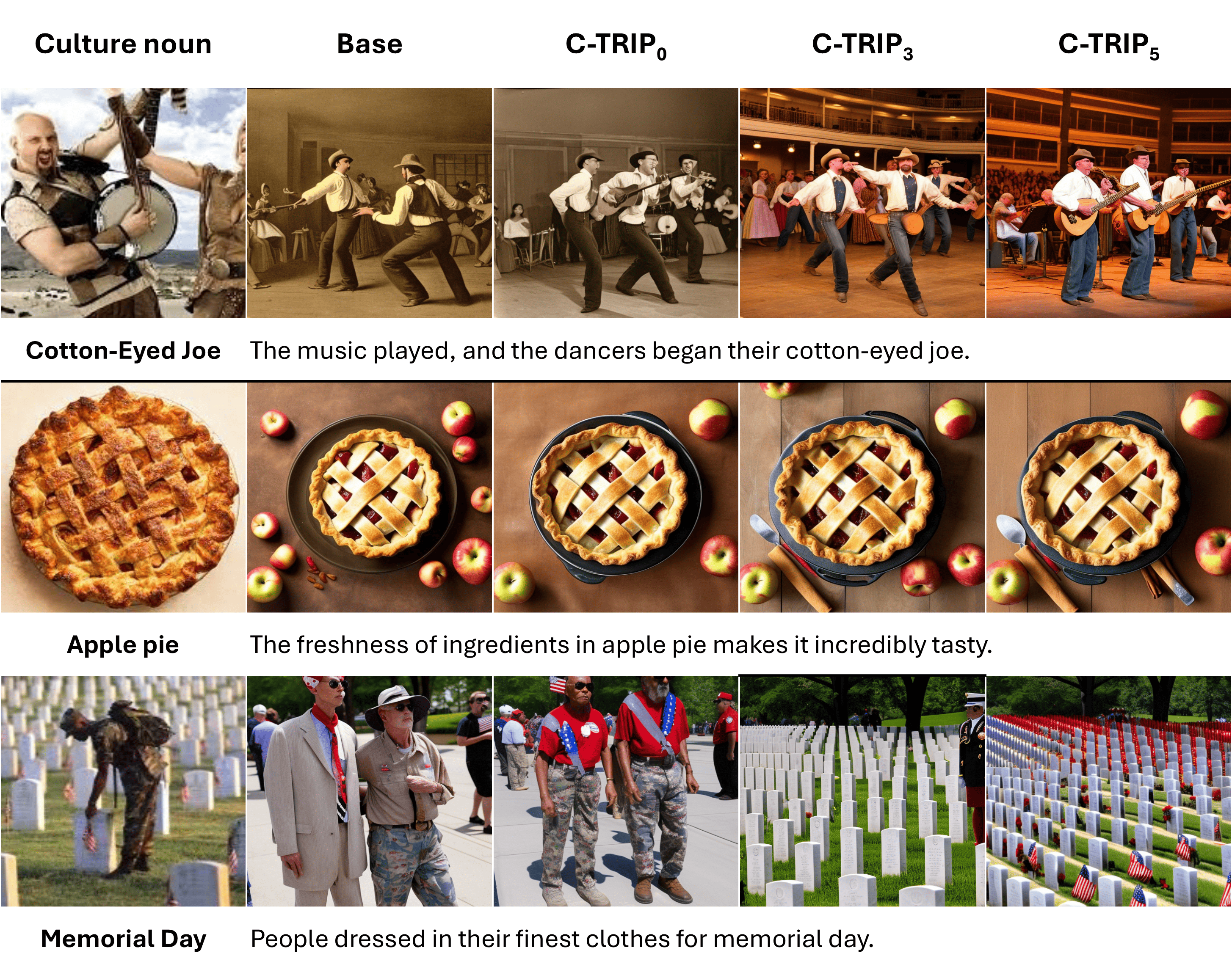}
    \caption{\textbf{Additional Qualitative Sample for American culture.} The top one image show the results of C-TRIP for culture nouns categorized under UC nouns (The United States has no culture nouns categorized under Q1.), while the bottom two images present the result for culture nouns categorized under RC nouns.}
    \label{UC_RC7}
\end{figure}

\begin{figure}
    \centering
    \includegraphics[width=\textwidth]{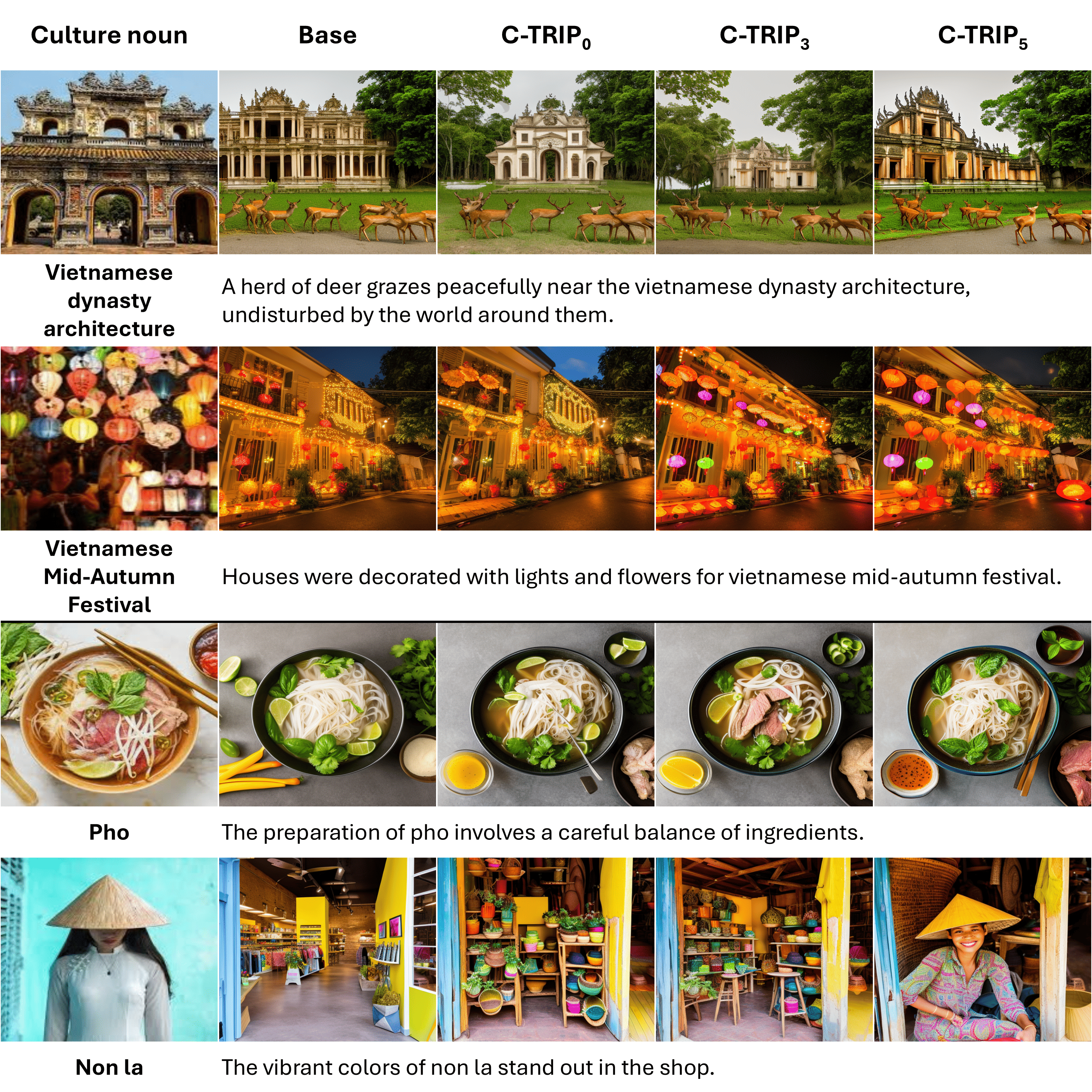}
    \caption{\textbf{Additional Qualitative Sample for Vietnamese culture.} The top two images show the results of C-TRIP for culture nouns categorized under UC nouns, while the bottom two images present the result for culture nouns categorized under RC nouns.}
    \label{UC_RC8}
\end{figure}

Furthermore, Figures \ref{Q1} to \ref{Q4} analyze the distribution of culture nouns across each quartile group. Looking at the top three countries in each group, for Q1, the leading countries were Germany, China, and Vietnam. For Q2, Japan, Vietnam, and India ranked highest. In Q3, China, India, and the USA were the top countries, while in Q4, the USA accounted for more than half, showing a dominant proportion.


\newpage

\begin{table}[H]
\resizebox{\columnwidth}{!}{%
\begin{tabular}{|c|ccc|cc|c|}
\hline
\multirow{2}{*}{} & \multicolumn{3}{c|}{Age}                                        & \multicolumn{2}{c|}{Gender}        & \multirow{2}{*}{Total} \\ \cline{2-6}
                  & \multicolumn{1}{c|}{21-30} & \multicolumn{1}{c|}{31-40} & 41-50 & \multicolumn{1}{c|}{Male} & Female &                        \\ \hline
China             & \multicolumn{1}{c|}{6}     & \multicolumn{1}{c|}{1}     & 0     & \multicolumn{1}{c|}{6}    & 1      & 7                      \\ \hline
Germany           & \multicolumn{1}{c|}{12}    & \multicolumn{1}{c|}{0}     & 0     & \multicolumn{1}{c|}{4}    & 8      & 12                     \\ \hline
India             & \multicolumn{1}{c|}{5}     & \multicolumn{1}{c|}{2}     & 0     & \multicolumn{1}{c|}{3}    & 4      & 7                      \\ \hline
Japan             & \multicolumn{1}{c|}{6}     & \multicolumn{1}{c|}{0}     & 0     & \multicolumn{1}{c|}{2}    & 4      & 6                      \\ \hline
Pakistan          & \multicolumn{1}{c|}{9}     & \multicolumn{1}{c|}{3}     & 0     & \multicolumn{1}{c|}{9}    & 3      & 12                     \\ \hline
South Korea       & \multicolumn{1}{c|}{9}     & \multicolumn{1}{c|}{1}     & 2     & \multicolumn{1}{c|}{9}    & 3      & 12                     \\ \hline
USA               & \multicolumn{1}{c|}{6}     & \multicolumn{1}{c|}{0}     & 0     & \multicolumn{1}{c|}{3}    & 3      & 6                      \\ \hline
Vietnam           & \multicolumn{1}{c|}{4}     & \multicolumn{1}{c|}{1}     & 0     & \multicolumn{1}{c|}{1}    & 4      & 5                      \\ \hline
\end{tabular}%
}
\caption{This table presents participant information, including age and gender distribution for each culture group.}
\label{Table3}
\end{table}

\begin{table}[H]
\resizebox{\columnwidth}{!}{%
\centering
\rowcolors{2}{gray!15}{white} 
\begin{tabular}{c cc cc}
\toprule
\multirow{2}{*}{Country} & \multicolumn{2}{c}{UC nouns} & \multicolumn{2}{c}{RC nouns} \\ \cmidrule(lr){2-3} \cmidrule(lr){4-5}
                         & Q1(\%)        & Q2(\%)        & Q3(\%)        & Q4(\%)       \\ \midrule
China                    & 12            & 24            & \textbf{40}   & 24           \\
Germany                  & \textbf{40}   & 12            & 28            & 20           \\
India                    & 16            & 24            & 28            & \textbf{32}  \\
Japan                    & 12            & \textbf{40}   & 28            & 20           \\
Pakistan                 & \textbf{40}   & 20            & 16            & 24           \\
South Korea              & \textbf{36}   & \textbf{36}   & 20            & 8            \\
USA                      & 0             & 8             & 28            & \textbf{64}  \\
Vietnam                  & \textbf{44}   & 36   & 12            & 8            \\ 
\bottomrule
\end{tabular}
}
\caption{Distribution of culture nouns by quartile group for each country. Bold text represents the highest proportion for each respective country.}
\label{CountryQProportion}
\end{table}

\newpage

\newpage
\onecolumn
\begin{table} [H]
\resizebox{\linewidth}{!}{%
\begin{tabularx}{\textwidth}{p{3.5cm}|X}
\hline
Category              & Culture Nouns                                                                                                                                                     \\ \hline \hline
architecture          & Tulou, Siheyuan, Chinese pagoda                                                                                                                                               \\ \hline
city \& landmark        & Dunhuang (Mogao Caves), Xi'an (Terracotta Army), Beijing (Forbidden City), Beijing (Temple of Heaven), Shanghai (Oriental Pearl Tower) \\ \hline
clothing              & Tangzhuang, Zhongshan suit, Qipao, Hanfu                                                                                                                                      \\ \hline
dance \& music           & Lion dance, Yangge                                                                                                                                                            \\ \hline
visual arts           & Jade art                                                                                                                                                                      \\ \hline
food \& drink        & Mooncake, Peking duck, Wonton, Xiaolongbao, Chinese hot pot                                                                                                                   \\ \hline
religion \& festival & Chinese New Year, Chinese Mid-Autumn Festival, Qingming                                                                                                                       \\ \hline
utensils \& tools    & Chinese wok, Chinese bamboo steamer                                                                                                                                           \\ \hline
\end{tabularx}
}
\caption{Categorization of culture nouns associated with Chinese culture.}
\label{tab5}
\end{table}

\begin{table}[H]
\begin{tabularx}{\textwidth}{p{3.5cm}|X}
\hline
Category              & Culture Nouns                                                                                                                                                                      \\ \hline \hline
architecture          & German Romanesque, German Gothic, German Baroque                                                                                                                                               \\ \hline
city \& landmark        & Berlin (Brandenburger Tor), Munich (Munich's Marienplatz), Cologne (Cologne Cathedral), Bavaria (Neuschwanstein Castle), Heidelberg (Heidelberg Castle) \\ \hline
clothing              & Lederhosen, Tracht, Dirndl, Schürze                                                                                                                                                            \\ \hline
dance \& music           & Zwiefacher, Schuhplattler                                                                                                                                                                      \\ \hline
visual arts           & Deutsch Renaissance                                                                                                                                                                            \\ \hline
food \& drink        & Mulled wine, Currywurst, Knödel, Bratwurst, Sauerkraut                                                                                                                                         \\ \hline
religion \& festival & Nikolaustag, Weihnachtsmärkte, Oktoberfest                                                                                                                                                     \\ \hline
utensils \& tools    & Bratpfanne, Rouladenklammer                                                                                                                                                                    \\ \hline
\end{tabularx}
\caption{Categorization of culture nouns associated with German culture.}
\label{tab6}
\end{table}

\begin{table}[H]
\begin{tabularx}{\textwidth}{p{3.5cm}|X}
\hline
Category              & Culture Nouns                                                                                                                                                     \\ \hline \hline
architecture          & Indian stupa, Mughal architecture, Gupta architecture                                                                                                                                               \\ \hline
city \& landmark        & Agra (Taj Mahal), Delhi (Red Fort), Jaipur (Amber Fort), Allahabad (Allahabad Fort), Delhi (Qutub Minar) \\ \hline
clothing              & Lehenga, Sari, Shalwar kameez, Dhoti                                                                                                                                      \\ \hline
dance \& music           & Manipuri dance, Bharatnatyam                                                                                                                                                            \\ \hline
visual arts           & Mughal painting                                                                                                                                                                      \\ \hline
food \& drink        & Paratha, Biryani, Vada Pav, Aloo Gobi, Saag                                                                                                                   \\ \hline
religion \& festival & Diwali, Holi, Durga Puja                                                                                                                       \\ \hline
utensils \& tools    & Tawa, Masala dabba                                                                                                                                           \\ \hline
\end{tabularx}
\caption{Categorization of culture nouns associated with Indian culture.}
\label{tab7}
\end{table}

\begin{table}[H]
\begin{tabularx}{\textwidth}{p{3.5cm}|X}
\hline
Category              & Culture Nouns                                                                                                                                                     \\ \hline \hline
architecture          & Shinto architecture, Torii, Edo architecture                                                                                                                                               \\ \hline
city \& landmark        & Tokyo (Tokyo Tower), Shizuoka (Mount Fuji), Kyoto (Kinkaku-ji), Osaka (Osaka Castle), Matsumoto (Matsumoto Castle) \\ \hline
clothing              & Hanten, Haori, Hakama, Yukata                                                                                                                                      \\ \hline
dance \& music           & Kabuki, Noh mai                                                                                                                                                            \\ \hline
visual arts           & Origami                                                                                                                                                                      \\ \hline
food \& drink        & Okonomiyaki, Tempura, Wagashi, Gyudon, Gyoza                                                                                                                   \\ \hline
religion \& festival & Sapporo Snow Festival, Gion Matsuri, Sanda Matsuri                                                                                                                       \\ \hline
utensils \& tools    & Sashimi bōchō, Takoyaki Pan                                                                                                                                           \\ \hline
\end{tabularx}
\caption{Categorization of culture nouns associated with Japanese culture.}
\label{tab8}
\end{table}

\begin{table}[H]
\begin{tabularx}{\textwidth}{p{3.5cm}|X}
\hline
Category              & Culture Nouns                                                                                                                                                     \\ \hline \hline
architecture          & Pakistani Buddhist architecture, Pakistani Indo-Islamic architecture, Pakistani Mughal architecture                                                                                                                                               \\ \hline
city \& landmark        & Lahore (Badshahi Mosque), Karachi (Mazar-e-Quaid), Islamabad (Faisal Mosque), Multan (Shrine of Bahauddin Zakariya), Hyderabad (Pakka Qila) \\ \hline
clothing              & Sherwani, Gharara, Shalwar Kameez, Lehenga                                                                                                                                      \\ \hline
dance \& music           & Khattak dance, Jhumair                                                                                                                                                            \\ \hline
visual arts           & Truck art                                                                                                                                                                      \\ \hline
food \& drink        & Biryani, Nihari, Gulab Jamun, Kheer, Gol Gappa                                                                                                                   \\ \hline
religion \& festival & Vaisakhi, Eid al-Fitr, Eid al-Adha                                                                                                                       \\ \hline
utensils \& tools    & Karahi, Degchi                                                                                                                                           \\ \hline
\end{tabularx}
\caption{Categorization of culture nouns associated with Pakistani culture.}
\label{tab9}
\end{table}

\begin{table}[H]
\begin{tabularx}{\textwidth}{p{3.5cm}|X}
\hline
Category              & Culture Nouns                                                                                                                                                     \\ \hline \hline
architecture          & Hanok, Korean pagoda, Korean temple                                                                                                                                           \\ \hline
city \& landmark        & Jeonju (Hanok Village), Seoul (Gyeongbokgung), Gyeongju (Bulguksa), Suwon (Hwaseong Fortress), Seoul (Jongmyo Shrine) \\ \hline
clothing              & Hanbok, Jeogori, Durumagi, Dangui                                                                                                                                              \\ \hline
dance \& music           & Cheoyongmu, Buchaechum                                                                                                                                                         \\ \hline
visual arts           & Minhwa                                                                                                                                                                         \\ \hline
food \& drink        & Bingsu, Kimchi, Sundubujjigae, Bibimbap, Tteokbokki                                                                                                                            \\ \hline
religion \& festival & Chuseok, Seollal, Korean New Year                                                                                                                                              \\ \hline
utensils \& tools    & Gamasot, Hangari                                                                                                                                                               \\ \hline
\end{tabularx}
\caption{Categorization of culture nouns associated with Korean culture.}
\label{tab10}
\end{table}

\begin{table}[H]
\begin{tabularx}{\textwidth}{p{3.5cm}|X}
\hline
Category              & Culture Nouns                                                                                                                                                     \\ \hline \hline
architecture          & Colonial Revival, Mission Revival, American Craftsman                                                                                                                                           \\ \hline
city \& landmark        & New York City (Statue of Liberty), San Francisco (Golden Gate Bridge), Washington (The White House), Chicago (Willis Tower), Los Angeles (Hollywood Sign) \\ \hline
clothing              & Cowboy hat, Denim overalls, Buffalo check shirt, Quilted vest                                                                                                                                              \\ \hline
dance \& music           & Bluegrass, Cotton-Eyed joe                                                                                                                                                         \\ \hline
visual arts           & American folk art                                                                                                                                                                         \\ \hline
food \& drink        & Apple pie, Buffalo wings, Clam chowder, Barbecue ribs, Cornbread                                                                                                                            \\ \hline
religion \& festival & Thanksgiving, Independence Day, Memorial Day                                                                                                                                              \\ \hline
utensils \& tools    & Cast iron skillet, Butter dish                                                                                                                                                               \\ \hline
\end{tabularx}
\caption{Categorization of culture nouns associated with  American culture.}
\label{tab11}
\end{table}

\begin{table}[H]
\begin{tabularx}{\textwidth}{p{3.5cm}|X}
\hline
Category              & Culture Nouns                                                                                                                                                     \\ \hline \hline
architecture          & Vietnamese dynasty architecture, Vietnamese stilt house, Vietnamese pagoda                                                                                                                                           \\ \hline
city \& landmark        & Hanoi (One Pillar Pagoda), Hanoi (Temple of Literature), Hanoi (Old Quarter), Hue (Imperial City of Hue), Quang Nam (My Son Sanctuary) \\ \hline
clothing              & Ao dai, Ao ba ba, Ao tu than, Non la                                                                                                                                              \\ \hline
dance \& music           & Mua lan, Quan ho                                                                                                                                                         \\ \hline
visual arts           & Vietnamese silk painting                                                                                                                                                                         \\ \hline
food \& drink        & Banh mi, Goi cuon, Pho, Bun cha, Banh xeo                                                                                                                            \\ \hline
religion \& festival & Vietnamese Lunar New Year, Vietnamese Mid-Autumn Festival, Hung Kings Temple Festival                                                                                                                                              \\ \hline
utensils \& tools    & Vietnamese wok, Vietnamese clay pot                                                                                                                                                               \\ \hline
\end{tabularx}
\caption{Categorization of culture nouns associated with Vietnamese culture.}
\label{tab12}
\end{table}


\begin{longtable}{cccp{0.5\textwidth}}
\hline
Country                   & Culture noun                  & \multicolumn{1}{c}{Configuration} & Prompt                                                                                                                                                                                                                                                                                                                                                                                                                                                                                                                                            \\ \hline
\multirow{4}{*}{Pakistan} & \multirow{4}{*}{Degchi}       & Base prompt               & The task was made easier with a degchi.                                                                                                                                                                                                                                                                                                                                                                                                                                                                                                           \\ \cline{3-4} 
                          &                               & C-TRIP\(_0\)              & The task was made easier with a degchi, a must-have utensil for any hardcore foodie, providing a seductive and smoked flavor to dishes.                                                                                                                                                                                                                                                                                                                                                                                                           \\ \cline{3-4} 
                          &                               & C-TRIP\(_3\)              & The task was made easier with a degchi, which allowed for efficient cooking and storage of ingredients.                                                                                                                                                                                                                                                                                                                                                                                                                                          \\ \cline{3-4} 
                          &                               & C-TRIP\(_5\)              & The task was made easier with a degchi, a type of traditional Indian cooking pot known for its rich flavors and aromas. This utensil was made from heavy-gauge brass and resembled a deep cooking vessel with enough space to cook and store milk, tea, vegetable curries, curd, and other ingredients. Its versatility in cooking both vegetarian and non-vegetarian dishes made it an essential tool.                                                                                                                                           \\ \hline
\multirow{4}{*}{China}    & \multirow{4}{*}{Jade art}     & Base prompt               & The gallery opening was a major event for jade art enthusiasts.                                                                                                                                                                                                                                                                                                                                                                                                                                                                                   \\ \cline{3-4} 
                          &                               & C-TRIP\(_0\)              & The gallery opening was a major event for jade art enthusiasts who appreciated the stone's subtle, translucent colors and its association with Chinese conceptions of the soul and immortality. The exhibit featured a range of jade pieces, from ancient artifacts to modern creations, showcasing the evolution of jade art over thousands of years. Attendees were treated to an immersive experience, surrounded by the beauty and symbolism of this revered stone.                                                                           \\ \cline{3-4} 
                          &                               & C-TRIP\(_3\)              & The gallery opening was a major event for jade art enthusiasts, who admired the subtle, translucent colors and artistic treatment of jade. The exhibit featured intricately crafted jade artifacts, created by skilled artisans over seven millennia ago. With gold, it represented the sky, earth, and four directions.                                                                                                                                                                                                                          \\ \cline{3-4} 
                          &                               & C-TRIP\(_5\)              & The gallery opening was a major event for jade art enthusiasts, who gathered to marvel at the exquisite carvings, jewelry, and ornaments on display, crafted from nephrite and jadeite minerals prized for their vibrant colors and high translucency. The exhibit featured pieces with delicate filigree patterns, ornate designs, and symbols of good fortune, showcasing the great skill and attention to detail involved in shaping these minerals into beautiful works of art. The event was a testament to the enduring appeal of jade art. \\ \hline
\multirow{4}{*}{Korea}    & \multirow{4}{*}{Hanbok}       & Base prompt               & hanbok are scattered on the floor after a long day.                                                                                                                                                                                                                                                                                                                                                                                                                                                                                               \\ \cline{3-4} 
                          &                               & C-TRIP\(_0\)              & hanbok are scattered on the floor after a long day, their delicate fabrics crumpled and disheveled. Despite the chaos, the vibrant colors and intricate patterns of the garments remain striking, a testament to the beauty and elegance of traditional Korean clothing.                                                                                                                                                                                                                                                                          \\ \cline{3-4} 
                          &                               & C-TRIP\(_3\)              & hanbok are scattered on the floor after a long day of festivities and celebrations. The hanbok is a traditional Korean dress that is worn for formal or semi-formal occasions. In modern times, Koreans often wear hanbok to cultural events and festivals.                                                                                                                                                                                                                                                                                       \\ \cline{3-4} 
                          &                               & C-TRIP\(_5\)              & Hanbok are scattered on the floor after a long day, their vibrant colors and intricate designs reflecting Korea's rich cultural heritage. The traditional clothing features a unique blend of straight and curved lines that are representative of a distinctly Korean aesthetic. With its roots tracing back to the Three Kingdoms of Korea period, these elegant dresses are typically worn on special occasions such as weddings and festivals.                                                                                                \\ \hline
\multirow{4}{*}{India}    & \multirow{4}{*}{Masala dabba} & Base prompt               & A simple masala dabba helped achieve the desired texture.                                                                                                                                                                                                                                                                                                                                                                                                                                                                                         \\ \cline{3-4} 
                          &                               & C-TRIP\(_0\)              & A simple masala dabba helped achieve the desired texture by providing quick access to the necessary spices without having to search through cabinets or drawers. The box's compact design made it easy to store on the countertop, keeping the workspace clutter-free and organized. With its small spoon, the cook could measure out the perfect amount of each spice.                                                                                                                                                                           \\ \cline{3-4} 
                          &                               & C-TRIP\(_3\)              & A simple masala dabba helped achieve the desired texture, by providing quick access to spices like cumin, coriander, and turmeric, which are essential for many Indian dishes.                                                                                                                                                                                                                                                                                                                                                                    \\ \cline{3-4} 
                          &                               & C-TRIP\(_5\)              & A simple masala dabba helped achieve the desired texture, its carefully arranged cups holding the secrets to the recipe's success. The cook knew that with this container, they could confidently combine the various spices and seasonings, resulting in a culinary masterpiece. It was an indispensable item in their kitchen, one that had been passed down through generations.                                                                                                                                                               \\ \hline
\multirow{4}{*}{USA}      & \multirow{4}{*}{Willis tower} & Base prompt               & chicago (willis tower) was beautifully lit up at night.                                                                                                                                                                                                                                                                                                                                                                                                                                                                                           \\ \cline{3-4} 
                          &                               & C-TRIP\(_0\)              & chicago (willis tower) was beautifully lit up at night, making it a stunning sight in the city's skyline. Its unique design and impressive height make it a popular spot for photographers and tourists alike. The building's facade is made of anodized aluminum and black glass, which adds to its visual appeal.                                                                                                                                                                                                                               \\ \cline{3-4} 
                          &                               & C-TRIP\(_3\)              & chicago (Willis Tower) was beautifully lit up at night, making it one of Chicago's most popular tourist destinations, with its facade made of anodized aluminum and black glass shining brightly.                                                                                                                                                                                                                                                                                                                                                 \\ \cline{3-4} 
                          &                               & C-TRIP\(_5\)              & The Willis Tower was beautifully lit up at night, its anodized aluminum and black glass façade shimmering against the city lights. The building's unique design took on a new level of sophistication in the evening hours, becoming a beacon of light in the Chicago skyline. Its nighttime beauty was a sight to behold.                                                                                                                                                                                                                        \\ \hline
\multirow{4}{*}{Vietnam}  & \multirow{4}{*}{Pho}          & Base prompt               & pho is a delicious way to start any meal.                                                                                                                                                                                                                                                                                                                                                                                                                                                                                                         \\ \cline{3-4} 
                          &                               & C-TRIP\(_0\)              & Pho is a delicious way to start any meal, with its warm, comforting broth and tender noodles providing a satisfying and filling experience.                                                                                                                                                                                                                                                                                                                                                                                                       \\ \cline{3-4} 
                          &                               & C-TRIP\(_3\)              & Pho is a delicious way to start any meal, with its warm, comforting broth and soft rice noodles providing a soothing contrast to the crunch of fresh herbs and the savory flavor of meat.                                                                                                                                                                                                                                                                                                                                                         \\ \cline{3-4} 
                          &                               & C-TRIP\(_5\)              & pho is a delicious way to start any meal, providing comfort and sustenance, whether as a light breakfast, a satisfying lunch, or a nourishing dinner.                                                                                                                                                                                                                                                                                                                                                                                             \\ \hline
\end{longtable}

\begin{table*}
    \centering
    \begin{tabularx}{\textwidth}{lX}
    \hline
         \textbf{Item Title}& \textbf{Item Text} \\
    \hline \hline
         Cultural Representation & Rank the images from 1 for the best representation of [country] culture to 4 for the worst cultural representation. Please ignore image artifacts (such as distorted faces, hands, or glitches) when considering Cultural Representation. \textbf{(1=most representative, 4=least representative)}\\ \hline
         The naturalness of the keyword & Keyword: [keyword], Rank the images from 1 for the most natural-looking to 4 for the least natural-looking. Please ignore image artifacts (such as distorted faces, hands, or glitches) when considering The naturalness of the keyword. \textbf{(1=most natural, 4=least natural)}\\ \hline
         Offensiveness & Rank the images by their offensiveness to you personally, numbering from 1 as the least offensive to 4 for the most offensive. Please ignore image artifacts (such as distorted faces, hands, or glitches) when considering Offensiveness. \textbf{(1=least offensive, 4=most offensive)}\\ \hline
         Description and Image Alignment & Image Description: [prompt], Rank the accuracy of the match between the image and its description. Number each image from 1 for the image best described by the description to 4 for the worst match between image and description. \textbf{(1=best described, 4=worst described)}\\ \hline
    \end{tabularx}
    \caption{The four survey items for our human evaluation survey.}
    \label{survey_item}
\end{table*}

\clearpage
\clearpage
\begin{longtable}{ccp{0.6\textwidth}}
\hline
\textbf{Country}                & \textbf{Quartile} & \textbf{Culture nouns (Count)}                                                                                                                                                                                                      \\ \hline \hline
\multirow{4}{*}{China}          & Q1                & Zhongshan suit (290), Siheyuan (408), Tangzhuang (462)                                                                                                                                                                                                               \\ \cline{2-3} 
                                & Q2                & Xiaolongbao (555), Dunhuang (Mogao Caves) (1042), Chinese bamboo steamer (1541), Chinese hot pot (2700), Tulou (2830), Qingming (2901)                                                                                                                                                 \\ \cline{2-3} 
                                & Q3                & Yangge (4115), Peking duck (5077), Shanghai (Oriental Pearl Tower) (5522), Chinese wok (5630), Chinese pagoda (10490), Chinese Mid-Autumn Festival (10645), Jade art (10856),  Xi'an (Terracotta Army) (12477), Mooncake (15494), Wonton (16147)                                                                                                     \\ \cline{2-3} 
                                & Q4                & Beijing (Temple of Heaven) (20108), Lion dance (27302), Qipao (42352), Beijing (Forbidden City) (55328), Hanfu (107289), Chinese New Year (391554)                                                                                                                       \\ \hline
\multirow{4}{*}{Germany}        & Q1                & Rouladenklammer (0), Zwiefacher (6), Deutsch Renaissance (12), Weihnachtsmärkte (17), Nikolaustag (66), Bratpfanne (113), Schuhplattler (129), German Romanesque (138), Knödel (409), Schürze (459)                                                                                                                \\ \cline{2-3} 
                                & Q2                & German Gothic (2105), German Baroque (2139), Currywurst (2831)                                                                                                                                                                                         \\ \cline{2-3} 
                                & Q3                & Tracht (3855), Heidelberg (Heidelberg Castle) (4885), Munich (Munich's Marienplatz) (5646), Bratwurst (7746), Lederhosen (8854), Berlin (Brandenburger Tor) (9286), Sauerkraut (17848)                                                                                                                    \\ \cline{2-3} 
                                & Q4                & Dirndl (19876), Bavaria (Neuschwanstein Castle) (21512), Cologne (Cologne Cathedral) (29416), Mulled wine (52917), Oktoberfest (111654)                                                                                                               \\ \hline
\multirow{4}{*}{India}          & Q1                & Gupta architecture (148), Allahabad (Allahabad Fort) (203), Indian stupa (298), Manipuri dance (321)                                                                                                                                                                                                            \\ \cline{2-3} 
                                & Q2                & Masala dabba (825), Bharatnatyam (1280), Vada pav (1355), Aloo gobi (1797), Mughal architecture (2736), Delhi (Qutub Minar) (3833)                                                                                                                                                                       \\ \cline{2-3} 
                                & Q3                & Mughal Painting (4618), Saag (5053), Shalwar kameez (8824), Tawa (8949), Jaipur (Amber Fort) (13909), Paratha (14272), Durga Puja (14904)                                                                                                                 \\ \cline{2-3} 
                                & Q4                & Dhoti (26053), Biryani (30699), Delhi (Red Fort) (43940), Agra (Taj Mahal) (92631), Holi (124249), Sari (129095), Diwali (207137), Lehenga (261160)                                                                                                                                              \\ \hline
\multirow{4}{*}{Japan}          & Q1                & Sashimi bōchō (0), Noh mai (15), Edo architecture (207)                                                                                                                                                                                           \\ \cline{2-3} 
                                & Q2                & Shinto architecture (505), Takoyaki pan (629), Gion Matsuri (711), Hanten (756), Gyudon (889), Sanja Matsuri (944), Sapporo Snow Festival (1603), Okonomiyaki (2976), Wagashi (3083), Matsumoto (Matsumoto Castle) (3242)                                                                                                                                                                               \\ \cline{2-3} 
                                & Q3                & Hakama (4957), Gyoza (6092), Haori (6293), Kyoto (Kinkaku-ji) (6591), Osaka (Osaka Castle) (7576), Yukata (14470), Tempura (15307)                                                                                                                                                            \\ \cline{2-3} 
                                & Q4                & Tokyo (Tokyo Tower) (21830), Torii (23570), Kabuki (27924), Shizuoka (Mount Fuji) (36331), Origami (566794)                                                                                                                           \\ \hline
\multirow{4}{*}{Pakistan}       & Q1                & Pakistani Indo-Islamic architecture (0), Pakistani Buddhist architecture (0), Jhumair (1), Hyderabad (Pakka Qila) (10), Multan (Shrine of Bahauddin Zakariya) (19), Pakistani Mughal architecture (29), Khattak dance (32), Degchi (50), Gol gappa (175), Karachi (Mazar-e-Quaid) (264)                                                                       \\ \cline{2-3} 
                                & Q2                & Nihari (733), Islamabad (Faisal mosque) (1591), Karahi (2327), Lahore (Badshahi Mosque) (2607), Vaisakhi (3046)                                                                                                                                                                                    \\ \cline{2-3} 
                                & Q3                & Gulab jamun (4181), Kheer (4285), Gharara (6853), Shalwar kameez (8824)                                                                                                                                                                  \\ \cline{2-3} 
                                & Q4                & Eid al-Fitr (17903), Eid al-Adha (18648), Biryani (30699), Sherwani (34183), Truck art (46479), Lehenga (261160)                                                                                                                                                                                                   \\ \hline
\multirow{4}{*}{South Korea}    & Q1                & Cheoyongmu (1), Gamasot (23), Durumagi (28), Aundubujjigae (29), Hangari (49), Buchaechum (49), Jeogori (200), Minhwa (310), Seollal (416)                                                                                                                                    \\ \cline{2-3} 
                                & Q2                & Seoul (Jongmyo Shrine) (523), Dangui (594), Korean pagoda (895), Bingsu (945), Gyeongju (Bulguksa) (1002), Tteokbokki (1026), Suwon (Hwaseong Fortress) (1491), Korean temple (1846), Jeonju (Hanok Village) (3018)                                                                                      \\ \cline{2-3} 
                                & Q3                & Chuseok (3853), Korean New Year (4840), Hanok (5797), Seoul (Gyeongbokgung) (7423), Bibimbap (7718)                                                                                                                                                                                         \\ \cline{2-3} 
                                & Q4                & Hanbok (30133), Kimchi (33340)                                                                                                                                                                                                                              \\ \hline
\multirow{4}{*}{USA}            & Q1                &                                                                                                                                                                                                                  \\ \cline{2-3} 
                                & Q2                & Cotton-Eyed Joe (542), Mission Revival (1909)                                                                                                                                                                                                                               \\ \cline{2-3} 
                                & Q3                & Buffalo check shirt (4342), American Craftsman (5101), Clam chowder (9757), Colonial Revival (11429), American folk art (11432), Barbecue ribs (11789), Chicago (Willis Tower) (12907)                                                                                           \\ \cline{2-3} 
                                & Q4                & Los Angeles (Hollywood Sign) (33952), Buffalo wings (34780), Cast iron skillet (41098), Denim overalls (43346), Butter dish (52346), Cornbread (55906), Quilted vest (61705), Bluegrass (96302), Cowboy hat (112205), San Francisco (Golden Gate Bridge) (120285), Apple pie (137320), New York City (Statue of Liberty) (180139), Memorial Day (270410), Independence Day (424057), Washington (The White House) (603026), Thanksgiving (1375806) \\ \hline
\multirow{4}{*}{Vietnam}        & Q1                & Vietnamese dynasty architecture (4), Vietnamese stilt house (15), Ao tu than (17), Vietnamese silk painting (21), Mua lan (36), Vietnamese wok (38), Ao ba ba (43), Vietnamese clay pot (68), Hung Kings Temple Festival (151), Vietnamese Mid-Autumn Festival (305), Goi cuon (385)                                                                                                            \\ \cline{2-3} 
                                & Q2                & Vietnamese pagoda (484), Hnoi (One Pillar Pagoda) (554), Banh Xeo (658), Quang Nam (My Son Sanctuary) (998), Quan ho (1132), Bun Cha (1301), Hue (Imperial City of Hue) (1814), Vietnamese Lunar New Year (3055), Hanoi (Temple of literature) (3628)                                                                                                  \\ \cline{2-3} 
                                & Q3                & Non la (5798), Banh mi (8550), Ao dai (12302)                                                                                                                                                                                 \\ \cline{2-3} 
                                & Q4                & Hanoi (Old Quarter) (26451), Pho (66696)                                                                                                                                                  \\ \hline
\caption{Culture noun counts by Quartile of each country}
\label{Table14}
\end{longtable}


\end{document}